\pdfoutput=1

\documentclass[11pt]{article}

\usepackage[preprint]{acl}

\usepackage{times}
\usepackage{latexsym}

\usepackage[T1]{fontenc}

\usepackage[utf8]{inputenc}

\usepackage{microtype}

\usepackage{inconsolata}

\usepackage{graphicx}

\usepackage{hyperref}
\usepackage{blindtext}
\usepackage{booktabs}
\usepackage{enumerate}
\usepackage{multirow}
\usepackage{enumitem}
\usepackage{multicol}
\usepackage{tabularray}
\usepackage{comment}
\usepackage{amsmath}
\usepackage{adjustbox}
\usepackage{placeins}
\usepackage{subcaption}

\usepackage{amssymb}
\usepackage{pifont}
\usepackage[dvipsnames]{xcolor}
\usepackage{colortbl}
\usepackage{pgfplots}
\usepackage{pgfplotstable}
\usepackage[most]{tcolorbox}
\usepackage{algorithm}
\usepackage{algpseudocode}
\usepackage{threeparttable}
\usepackage[normalem]{ulem}
\usepackage{geometry}
\usepackage{tikz}
\usepackage{tabularx}

\pgfplotsset{compat=1.18}
\usepgfplotslibrary{groupplots}
\definecolor{lightbrown}{RGB}{243, 236, 226}
\definecolor{colorBlue}{HTML}{378ADD}
\definecolor{colorTeal}{HTML}{1D9E75}
\definecolor{colorOrange}{HTML}{EF9F27}
\definecolor{colorRed}{HTML}{E24B4A}
\definecolor{colorUnrescued}{HTML}{D85A30}
\usepackage{listings}
\lstdefinestyle{sqlappendix}{
  language=SQL,
  basicstyle=\ttfamily\footnotesize,
  keywordstyle=\bfseries,
  commentstyle=\itshape,
  showstringspaces=false,
  breaklines=true,
  breakatwhitespace=false,
  columns=fullflexible,
  keepspaces=true,
  upquote=true,
  frame=single,
  rulecolor=\color{black!25},
  framerule=0.4pt,
  framesep=3pt,
  xleftmargin=0pt,
  xrightmargin=0pt,
  aboveskip=4pt,
  belowskip=6pt,
  escapeinside={(*@}{@*)}
}
\algrenewcommand\algorithmicindent{0.8em}

\title{{GRIPS}:  An Empirical Study of Pre-SQL Grounding
Information for SQL Generation}

\author{Ekata Mitra, Yufei Tao, Kamalamalini Nagasundaram \and Ameeta Agrawal\\
  Department of Computer Science, Portland State University, USA \\
  \texttt{\{ekata, yutao,kamalan,ameeta\}@pdx.edu}}

\begin{document}
\maketitle
\begin{abstract}

Text-to-SQL systems answer natural language questions over structured data, but
can fail before SQL generation by linking question mentions to the wrong table,
column, value, or relation. Even valid SQL can then return an incorrect answer. This raises a central question:
\textit{What pre-SQL information helps a SQL generator, and when does
additional information cease to help?} We investigate this question on
TempTabQA-C and BIRD. Among the audited SQL Static failures, grounding errors form the largest
category on both benchmarks. We introduce \textsc{GRIPS} (Grounding Representation and Input Processing for
SQL generation), a three-stage method that derives structured pre-SQL
information to make these decisions explicit before query generation.~Across GPT-5-nano and Llama-70B, GRIPS-Adaptive improves over SQL Static in
18 of 20 settings, with an average accuracy gain of 11.8 percentage points,
while reducing model-call token usage by 47.8\%. Our controlled
analysis shows which
grounding information can improve SQL generation, whereas adding more pre-SQL
information does not consistently help. By separating question--input grounding from SQL
generation, \textsc{GRIPS} makes text-to-SQL more accurate, efficient, and
easier to diagnose. 
\end{abstract}

\section{Introduction}
Text-to-SQL systems provide a practical interface for answering natural language questions over structured data: given a schema or set of tables, the system returns an executable query and a final answer. Large language models (LLMs) have made this workflow look straightforward, but structured-data reasoning requires several decisions before any SQL is generated. The system must determine what the question refers to in the schema, select relevant tables or slices, resolve values and filters, and decide which operations to apply. These pre-SQL decisions are
interdependent: an incorrect grounding decision can lead to a valid query that
answers the wrong question. Selecting relevant context is particularly important
because more context is not necessarily better; even when retrieved information
is relevant, additional context can degrade LLM performance
\citep{du-etal-2025-context}. Errors in these pre-SQL decisions propagate to downstream reasoning, and have been observed across semi-structured tables and fully relational databases \citep{zhou2024freb,
shankarampeta-etal-2025-transienttables,
gruber-etal-2025-complextempqa,
li-etal-2023-bird,
oses-grijalba-etal-2025-semeval-task8}.

Prior work improves different parts of the text-to-SQL pipeline through schema
linking and retrieval
\citep{gao2024dail,
shi2025gen,
glass-etal-2025-extractive-schema-linking,
nahid-etal-2025-rethinking-schema-linking,
eben-etal-2025-rasl,
pourreza2023dinsqldecomposedincontextlearning,
wang2025macsqlmultiagentcollaborativeframework,
xie2026sde,
talaei2024chesscontextualharnessingefficient,
rajgaria-etal-2025-universal},
cost-aware routing
\citep{zhu2025elliesql,yi2025sktsql,shao2025enhancing},
and tabular QA prompting
\citep{kulkarni2025llm,
deng-etal-2025-enhancing,
rajgaria-etal-2025-universal}.
However, the
question--input interpretation must be systematically examined
while keeping the SQL generator fixed. Consequently, it remains difficult to
isolate which pre-SQL information helps and when additional
information hurts.

To examine this issue, we use SQL Static \citep{kulkarni2025llm} as a common diagnostic system  across two datasets, TempTabQA-C and BIRD. Our audit shows that many failures originate in question--input grounding
before SQL syntax or execution becomes the main issue. This observation motivates treating pre-SQL
interpretation as an explicit, inspectable handoff rather than leaving it only
to the SQL-generation prompt. Therefore, we introduce \textsc{GRIPS} (\textit{Grounding Representation and Input Processing for SQL generation}), a three-stage architecture  that separates input processing, SQL generation, and
execution. Stage~I derives structured
pre-SQL information from the question and its available context. Stage~II
combines this information with the original input and SQL demonstrations to
generate a single SQLite query. Stage~III executes the query and returns the
answer. We study a fixed-prompt version, \textsc{GRIPS}-Static, and an
adaptive version, \textsc{GRIPS}-Adaptive, which selects demonstrations for
the current input. This separation allows us to study not only
whether GRIPS improves SQL generation, but also which information should be
resolved before generation begins.

Our work is organized around three research questions: 

\noindent \textbf{RQ1:} Where do text-to-SQL failures arise---during
question--input grounding, SQL operation selection, or execution? 

\noindent \textbf{RQ2:}  How does the complete \textsc{GRIPS} system compare
with prior methods in answer accuracy and model-call token usage?

\noindent \textbf{RQ3:} When the SQL generator and demonstrations are fixed,
which pre-SQL grounding information improves performance?

Our main contributions are:
\begin{enumerate}[leftmargin=*,itemsep=2pt,topsep=2pt]
    \item A cross-format analysis of text-to-SQL failures, with a diagnostic
    taxonomy covering grounding, operator, and execution-interface errors.

    \item An inspectable pre-SQL interface, evaluated through the
    \textsc{GRIPS} pipeline across two structured-data settings and two
    model families.

    \item A controlled study of the information passed through this
    interface, showing that compact grounding can help while additional
    fields do not consistently improve SQL generation.
\end{enumerate}


\section{Related Work}

\paragraph{Schema linking.}
A key challenge in Text-to-SQL systems is schema linking. Recent methods improve this step through schema retrieval, lexical and semantic matching, pseudo-schema construction, and context filtering \citep{gao2024dail, shi2025gen, glass-etal-2025-extractive-schema-linking, nahid-etal-2025-rethinking-schema-linking, eben-etal-2025-rasl, talaei2024chesscontextualharnessingefficient}. Other approaches decompose generation into schema selection, query generation, validation, correction, or database exploration \citep{pourreza2023dinsqldecomposedincontextlearning, wang2025macsqlmultiagentcollaborativeframework, xie2026sde}.

\begin{table}[!t]
\centering
\small
\setlength{\tabcolsep}{4pt}
\begin{tabularx}{\columnwidth}{@{}lX}
\toprule
\multicolumn{2}{@{}l}{\textbf{Example 1: TempTabQA-C, Hard}}\\
\midrule
C: & Darrell Pace \\
   & Personal Information: Birth Date = 1956-10-23 \\
   & Olympic Games: Individual \quad MedalGold $\mid$ 1976 $\mid$ Montreal; MedalGold $\mid$ 1984 $\mid$ Los Angeles \\
   & Pan American Games: Team \quad MedalGold $\mid$ 1979 $\mid$ San Juan; MedalGold $\mid$ 1983 $\mid$ Caracas \\
   & Pan American Games: Individual \quad MedalGold $\mid$ 1983 $\mid$ Caracas; MedalSilver $\mid$ 1979 $\mid$ San Juan \\
Q: & How many medals did Darrell Pace win in his/her twenties? \\
A: & 6 \\
\midrule
\multicolumn{2}{@{}l}{\textbf{Example 2: BIRD-Mini-Dev, \textit{california\_schools}}}\\
\midrule
E: & ``between 1/1/1980 and 12/31/1980'' means the year = 1980. \\
Q: & What is the average number of test takers from Fresno schools that opened between 1/1/1980 and 12/31/1980? \\
A: & 137.89 \\
\bottomrule
\end{tabularx}
\caption{Examples from TempTabQA-C and BIRD where C: Table Context; Q: Question; A:
Answer; E: Evidence.}
\label{tab:qualitative-dataset-examples}
\end{table}

\paragraph{Prompting method.}
Few-shot prompting, decomposition, routing, and retrieval are commonly used to improve SQL generation quality while controlling LLM inference cost \citep{pourreza2023dinsqldecomposedincontextlearning, zhu2025elliesql, yi2025sktsql, shao2025enhancing}. Adaptive prompting further selects examples or strategies based on properties of the current question \citep{rajgaria-etal-2025-universal}. Our Adaptive configuration follows this general direction while the controlled experiments isolate the effect of the structured input record itself.

\paragraph{Symbolic method.}
Recent work combines LLM prompting, symbolic SQL generation, and execution for temporal table reasoning \citep{gupta-etal-2023-temptabqa, deng-etal-2025-enhancing,kulkarni2025llm}. Robustness studies further show that table reasoning remains sensitive to changes in question form, table structure, and context \citep{zhou2024freb, shankarampeta-etal-2025-transienttables, gruber-etal-2025-complextempqa}. Our approach complements this work by making pre-SQL question--input matches explicit.


\section{Diagnosing Pre-SQL Failures in Structured-Data Reasoning}

We analyze SQL Static \citep{kulkarni2025llm}, because its fixed-prompt,
single-query design enables direct failure analysis, on TempTabQA-C and BIRD.
We group errors into question--input misalignment, incorrect SQL operations,
and execution failures.

\subsection{Benchmark}
Our analysis spans two structurally different benchmarks, with examples shown in Table~\ref{tab:qualitative-dataset-examples}.

\noindent \textbf{TempTabQA-C} \citep{gupta-etal-2023-temptabqa} is a 
temporal table QA benchmark built from Olympic and athletics 
infoboxes. It provides seven controlled evaluation splits: three 
difficulty levels (\textit{Easy} (732), \textit{Medium} (507), \textit{Hard} (719)), 
the original unmodified split (\textit{Orig.} (578)), a counterfactual 
variant with perturbed table content (\textit{CFact.} (699)), and two 
table-size variants (\textit{Small} (855), \textit{Large} (538)). Together 
these splits test robustness across question difficulty, 
distributional shift, and structural variation.

\noindent \textbf{BIRD} \citep{li-etal-2023-bird} is a 
text-to-SQL benchmark spanning eleven real-world relational databases. For answer-level evaluation, we execute the human-authored gold SQL queries to derive the reference outputs.
Each question has a gold SQL query, an evidence hint, and one of three hardness labels: 
\textit{Simple} (92), \textit{Moderate} (92), or \textit{Challenging} (92). {We use the Mini-Dev split throughout and refer to it as {BIRD}.}

\subsection{Analysis}
We audit error cases from SQL Static predictions with incorrect answers.~For TempTabQA-C, we use the Easy, Medium, and Hard splits to match BIRD's three difficulty groups:
Simple, Moderate, and Challenging. SQL Static gives 484
incorrect answers out of 1,958 TempTabQA-C questions and 153 out of
276 BIRD questions. Following the error patterns described by \citet{kulkarni2025llm}, we assign each failure to one of these three classes:

\begin{itemize}[leftmargin=*,itemsep=2pt,topsep=2pt]
    \item \textbf{Grounding (G):} the question is linked to the wrong field, value, filter, table slice, or join path.
    \item \textbf{Operator (O):} the relevant information is identified, but the SQL uses the wrong temporal condition, aggregation, ranking, comparison, or selection logic.
    \item \textbf{Execution-Interface (E):} the system does not produce a usable query or answer, for example because the SQL is invalid, empty, or malformed.
\end{itemize}
For each failure, we inspect the question, permitted context or schema,
evidence, generated SQL, execution result, and reference answer, and
assign one primary label according to where the error first appears:
before computation (G), during computation (O), or at the execution
interface (E). 

The complete annotation protocol and nine-way taxonomy are provided in
Appendix~\ref{sec:appendix_failure_analysis} along with one example for
each observed subtype.

\begin{figure}[!t]
    \centering
    \includegraphics[width=\linewidth]{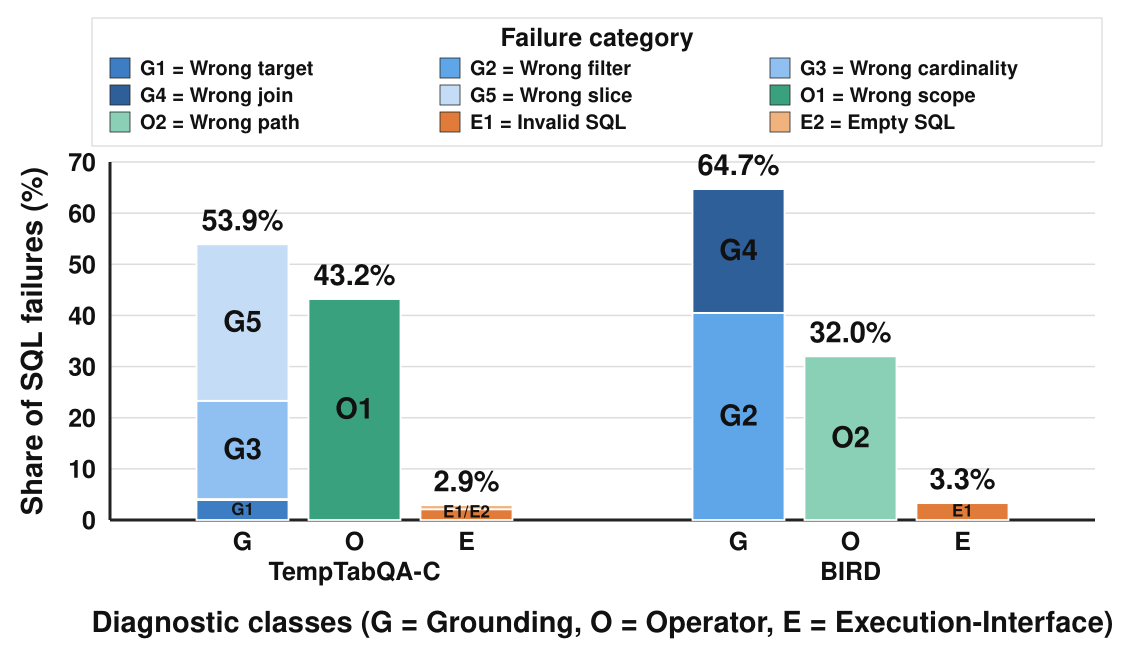}
    \caption{SQL Static failure breakdown by diagnostic class 
and subcategory. TempTabQA-C grounding errors concentrate 
on schema-slice and cardinality (G5/G3); BIRD on filter 
and join grounding (G2/G4).}
    \label{fig:failure-composition}
\end{figure}

\begin{table}[!t]
\small
\centering
\begin{adjustbox}{max width=\linewidth}
\begin{tabular}{@{}p{0.12\linewidth} p{0.41\linewidth} p{0.32\linewidth}@{}}
\toprule
\textbf{Type} & \textbf{Example Question} & \textbf{Root Cause} \\
\midrule
Wrong target (G) &
\textit{Which was the most current medal win for Lin Sang?} &
Returns a partial event string instead of the correct format; target field never resolved. \\
\addlinespace
Wrong scope (O) &
\textit{How many times did Ilya Shymanovich win an International 
medal in the European Championships (SC) before 2019?} &
Schema slice identified but temporal operator applied 
incorrectly; produces 0 instead of gold count 4. \\
\addlinespace
Empty SQL (E) &
\textit{What is the shortest time span (in years) within which 
Abhinav Bindra won gold, silver, and bronze medals in the same 
format across any tournament?} &
Empty SQL produced; pipeline fails before any executable query 
is formed. \\
\bottomrule
\end{tabular}
\end{adjustbox}
\caption{One representative failure per diagnostic class from 
SQL Static.}
\label{tab:failure_examples}
\end{table}

\begin{figure*}[!t]
\centering
\includegraphics[width=1\textwidth]{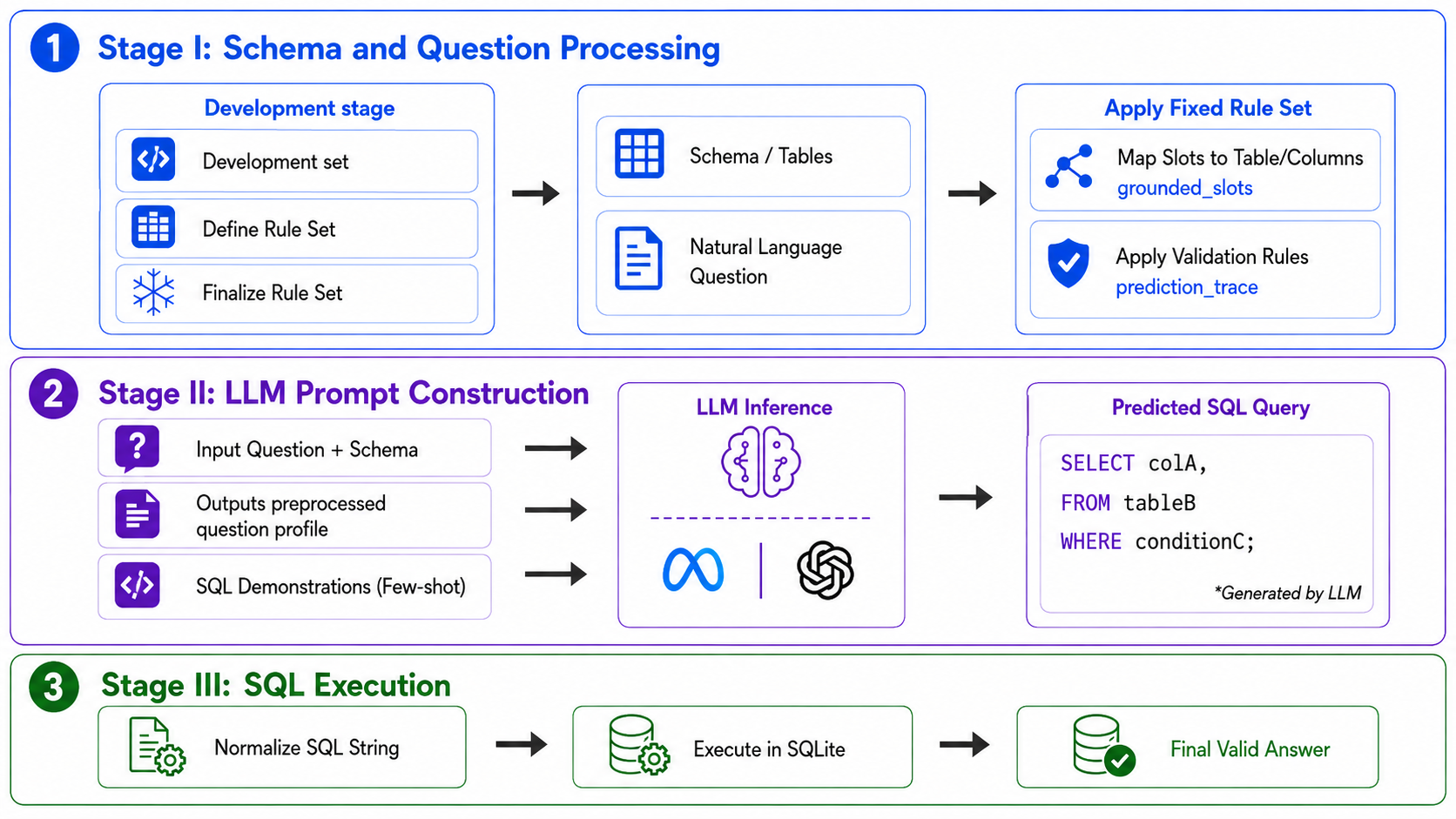}
\caption{Overview of \textsc{GRIPS} pipeline. Stage~I maps the question and context to a structured pre-SQL record using a frozen benchmark-specific rule-set. Stage~II combines the original input, the Stage~I record, and SQL demonstrations to generate SQL query. Stage~III normalizes and executes the query to produce the final answer.}
\label{fig:method-pipeline}
\end{figure*}

\subsection{RQ1: Where does structured-data reasoning fail?}

As shown in Figure~\ref{fig:failure-composition}, grounding is the largest
diagnostic class across both benchmarks. It accounts for 53.9\% of analyzed
TempTabQA-C errors and 64.7\% of BIRD errors. Operator errors are
next largest, accounting for 43.2\% and 32.0\%, whereas execution-interface
errors are rare (2.9\% and 3.3\%). 

Table~\ref{tab:failure_examples} gives
representative cases. Several outputs contain valid SQL but still return the
wrong answer because they use an incorrect target, filter, table slice, or
join path. 
These findings motivate \textsc{GRIPS}: rather than leaving
question--input connections implicit in the prompt, Stage~I makes them
explicit before SQL generation. %

\section{The GRIPS Framework}
\label{sec:method}
We introduce GRIPS, a three-stage framework for structured-data reasoning.  Figure~\ref{fig:method-pipeline} illustrates the workflow.

Given a question and its permitted table or schema context, GRIPS separates the
task into (1) structured input processing, (2) LLM-based SQL generation, and
(3) query execution. Rather than requiring the LLM to infer relevant fields,
values, filters, tables, and relations while generating SQL, Stage~I derives a
compact structured record from the question and available context. Stage~II
uses this record together with the original input and demonstrations to generate
a query, which Stage~III executes to return the final answer. 

{Stage I targets grounding errors by making schema and value matches explicit; Stage II targets operator errors by conditioning SQL generation on this record and on matching demonstrations; Stage III executes the query and exposes invalid or empty outputs.}

\begin{table*}[!t]
\centering

\centering

\begin{minipage}[t]{0.485\textwidth}
\vspace{0pt}
\begin{tcolorbox}[
    enhanced,
    equal height group=stage-one-b,
    colback=lightbrown,
    colframe=black!50!brown,
    colbacktitle=black!50!brown,
    coltitle=white,
    title={Example 1: TempTabQA-C, Hard},
    fonttitle=\bfseries\small,
    fontupper=\small,
    boxrule=0.7pt,
    arc=1.5mm,
    left=1.5mm,
    right=1.5mm,
    top=1mm,
    bottom=1mm,
    boxsep=0.5mm,
    before upper={\setlength{\parindent}{0pt}\raggedright}
]

{\bfseries Current input}\par
\hspace*{1.2em}Question and supplied infobox: Table~\ref{tab:qualitative-dataset-examples}, Example~1.\\[3pt]

{\bfseries Frozen rule-set configuration}\par
\hspace*{1.2em}\texttt{SLOT\_ORDER=[Athlete,Tournament,}\\
\hspace*{2.4em}\texttt{Format,MedalColor,Year]}\\
\hspace*{1.2em}Fixed normalization and medal-record filtering rules.\\[3pt]

{\bfseries Deterministic processing}\par
\hspace*{1.2em}Normalize the input, identify possible matches, resolve overlapping matches, remove duplicates, and compute the candidate-record profile.\\[3pt]

{\bfseries Stage~I JSON output}\par
\hspace*{1.2em}\texttt{"grounded\_slots": \{}\\
\hspace*{2.4em}\texttt{"Athlete": "Darrell Pace"}\\
\hspace*{1.2em}\texttt{\},}\\[2pt]

\hspace*{1.2em}\texttt{"prediction\_trace": \{}\\
\hspace*{2.4em}\texttt{"record\_count": 11,}\\
\hspace*{2.4em}\texttt{"candidate\_record\_count": 11,}\\
\hspace*{2.4em}\texttt{"slot\_priority": "anchor\_only",}\\
\hspace*{2.4em}\texttt{"prediction": "Hard"}\\
\hspace*{1.2em}\texttt{\}}

\end{tcolorbox}
\end{minipage}
\hfill
\begin{minipage}[t]{0.485\textwidth}
\vspace{0pt}
\begin{tcolorbox}[
    enhanced,
    equal height group=stage-one-b,
    colback=lightbrown,
    colframe=black!50!brown,
    colbacktitle=black!50!brown,
    coltitle=white,
    title={Example 2: BIRD (\textit{california\_schools})},
    fonttitle=\bfseries\small,
    fontupper=\small,
    boxrule=0.7pt,
    arc=1.5mm,
    left=1.5mm,
    right=1.5mm,
    top=1mm,
    bottom=1mm,
    boxsep=0.5mm,
    before upper={\setlength{\parindent}{0pt}\raggedright}
]

{\bfseries Current input}\par
\hspace*{1.2em}Question, database schema, and benchmark-provided evidence: Table~\ref{tab:qualitative-dataset-examples}, Example~2.\\[3pt]

{\bfseries Frozen rule-set configuration}\par
\hspace*{1.2em}\texttt{SLOT\_ORDER=[Table,Column,}\\
\hspace*{2.4em}\texttt{Value,Date]}\\
\hspace*{1.2em}\texttt{grounding\_uses\_evidence=true}\\
\hspace*{1.2em}Fixed schema-normalization and foreign-key rules.\\[3pt]

{\bfseries Deterministic processing}\par
\hspace*{1.2em}Normalize the input, identify possible question--schema and question--evidence matches, resolve overlaps, remove duplicates, and compute the relevant schema profile.\\[3pt]

{\bfseries Stage~I JSON output}\par
\hspace*{1.2em}\texttt{"grounded\_slots": \{}\\
\hspace*{2.4em}\texttt{"Column": ["Number of Test Takers",}\\
\hspace*{3.6em}\texttt{"School"],}\\
\hspace*{2.4em}\texttt{"Date": ["1/1/1980", "12/31/1980"],}\\
\hspace*{2.4em}\texttt{"Value": ["1", "1980"]}\\
\hspace*{1.2em}\texttt{\},}\\[2pt]

\hspace*{1.2em}\texttt{"prediction\_trace": \{}\\
\hspace*{2.4em}\texttt{"implicated\_tables": ["schools"],}\\
\hspace*{2.4em}\texttt{"shortest\_fk\_path": 0,}\\
\hspace*{2.4em}\texttt{"prediction": "moderate"}\\
\hspace*{1.2em}\texttt{\}}

\end{tcolorbox}
\end{minipage}

\caption{Stage~I outputs for the running examples in
Table~\ref{tab:qualitative-dataset-examples}.}
\label{tab:stage1-examples}
\end{table*}
\subsection{Stage I: Schema and Question Processing}
\label{sec: stage I}
\paragraph{Setup.}

Stage~I has two phases: development-time rule-set construction and evaluation-time input processing.  In the development step, we use a development subset from each benchmark to determine which question mentions and structural properties to record, establish the rule order, and set a small number of fixed thresholds. We then finalize the benchmark-specific rule set and apply it unchanged to all remaining evaluation examples.

For TempTabQA-C, the development set contains 120 examples: 40 from each of
the Easy, Medium, and Hard splits. Each split includes 20 SQL Static successes
and 20 failures, covering variation in question mentions, event filters,
temporal conditions, and candidate table records. For BIRD, the development
set contains 30 examples: 10 examples from each gold difficulty
level---Simple, Moderate, and Challenging---including 5 successes and 5
failures per level. These examples cover both direct schema access and more
constrained multi-table questions, allowing us to examine connections that SQL
Static handles correctly as well as those associated with incorrect SQL. Development set examples are excluded from the corresponding evaluation runs.

\begin{table}[!t]
\centering
\footnotesize
\setlength{\tabcolsep}{3.2pt}
\begin{adjustbox}{max width=\linewidth}
\begin{tabular}{p{0.27\columnwidth}p{0.65\columnwidth}}
\toprule
\textbf{Layer} & \textbf{Contents} \\
\midrule
Reusable operations & normalization, span detection, overlap resolution,
counting, context filtering, graph traversal, JSON serialization \\
Benchmark configuration & slot labels, lexical cues and aliases, context
grammar, schema conventions, fixed thresholds \\
Per-example inputs & question mentions, supplied schema or table context,
benchmark evidence \\
\bottomrule
\end{tabular}
\end{adjustbox}
\caption{Components of the Stage~I preprocessing configuration.}
\label{tab:stage1-boundary}
\end{table}
\paragraph{Constructing the rule set.}

As TempTabQA-C and BIRD are substantially different in structure, we use a separate
Stage~I rule-set for each benchmark. We use the development sets to refine slot maps, rule order, and fixed thresholds used by the structural profile. TempTabQA-C contains semi-structured
temporal tables, while BIRD uses relational database schemas. Each rule-set converts the question and its available context into the same compact JSON format.

To select these rules, we first map each question to its table or
schema context, and to benchmark-provided evidence for BIRD. The recognized
matches are stored in \texttt{grounded\_slots}. Their coverage indicates how
much of the question is directly supported by the input: an anchor-only match
leaves more information to infer before SQL generation. We combine this signal
with candidate-record counts, filter counts, implicated tables, and foreign-key
distance to assign a coarse \texttt{prediction\_trace} label.

Following the general idea of predicting question difficulty or hardness in Spider
\citep{yu2018spider}, we derive this \texttt{prediction\_trace} label from question--input mappings rather
than gold SQL. For TempTabQA-C, we vary the candidate-record cutoff on
development examples and select five because it gives the strongest agreement
with the tested difficulty labels. For BIRD, we check simple boundaries,
including foreign-key distance and the number of implicated tables, only on development
examples. We then freeze the slot maps, rule order, and thresholds for all
evaluation runs.

Tables~\ref{tab:stage1-examples} and~\ref{tab:stage1-boundary} illustrate the
Stage~I records and distinguish reusable processing from benchmark-specific
rules. Appendix~\ref{sec:appendix-stage1} gives the complete Stage~I
construction, including additional examples, structural fields, and
development-set threshold checks.

\paragraph{Evaluation-time input processing.}
At evaluation time, the frozen rule-set normalizes the input--question and context, produces two JSON fields:
\texttt{grounded\_slots}, which lists the recognized matches, and
\texttt{prediction\_trace}, which summarizes simple properties of the current
input. For TempTabQA-C these properties include candidate medal-record
counts. For BIRD they include the involved tables and foreign-key distance.

\subsection{Stage II: LLM Prompt Construction and SQL Generation}

Stage~II provides the LLM with the original benchmark input and the Stage~I JSON record. The prompt contains:
\begin{itemize}[leftmargin=*,itemsep=2pt,topsep=2pt]
    \item an instruction to return exactly one executable SQLite \texttt{SELECT} query;
    \item the original question;
    \item the context or database schema;
    \item benchmark-provided evidence, when available(BIRD);
    \item the Stage~I~JSON record containing \texttt{grounded\_slots} and \texttt{prediction\_trace};
    \item six SQL demonstrations.
\end{itemize}

The demonstrations are benchmark-specific but identical across the two model families. They show the required output format and examples of mapping natural-language questions to executable SQLite queries.

We evaluate two variants. 

\noindent (1) \textbf{GRIPS-Static} uses the Stage~I JSON with a fixed set of demonstrations, isolating the effect of explicit grounding.

\noindent (2) \textbf{GRIPS-Adaptive} also includes the Stage~I JSON record, but uses additional Stage~I structural fields to dynamically select demonstrations. 

Appendix~\ref{sec:appendix-stage2} gives the full prompt and
demonstration-selection details.

\subsection{Stage III: SQL Execution}

Stage~III normalizes the SQL generated by the LLM and executes it in SQLite. The execution result, rather than the raw LLM response, is used as the final answer.

We use the same execution procedure for all methods. This keeps answer recovery consistent across systems and helps identify whether an error is associated with Stage~I processing, demonstration selection, SQL generation, or query execution.

\begin{table*}[!t]
\centering
\small
\setlength{\tabcolsep}{3.8pt}
\begin{adjustbox}{max width=\linewidth}
\begin{tabular}{llcccccccccc}
\toprule
\multirow{2}{*}{Model}
& \multirow{2}{*}{Method}
& \multicolumn{7}{c}{\textbf{TempTabQA-C} (Accuracy $\uparrow$)}
& \multicolumn{3}{c}{\textbf{BIRD} (Accuracy $\uparrow$)} \\
\cmidrule(lr){3-9} \cmidrule(lr){10-12}
&
& Easy & Medium & Hard & Orig. & CFact. & Small & Large
& Simple & Moderate & Chall. \\
\midrule
GPT-5-nano
 & Direct Static  & 90.57 & 61.54 & 55.22 & 69.55 & 57.31  & 78.71 & 58.74
& 6.52 & 3.26 & 2.17 \\

& SQL Static
& 77.73 & 63.31 & 60.64 & 64.71 & 60.20 & 75.79 & 66.54
& \textbf{58.70} & 34.78 & 21.74 \\
 & DIN-SQL
& 69.95 & 35.70 & 43.53 & 47.06
& 41.49 & 59.18 & 49.07 & 48.91 & 30.43 & 21.74 \\
& MAC-SQL
& 83.88 & 59.96 & 51.32 & 66.26 & 60.52 & 73.10 & 65.61
& 44.57 & 26.09 & 25.00 \\
\arrayrulecolor{gray}\midrule 
Ours
 & GRIPS-Static   & 84.29 & 75.54 & 62.17 & 76.64 & 66.38 & 82.11 & 73.42
& 54.35 & 33.70 & {27.17} \\
& GRIPS Adaptive
& \textbf{92.35} & \textbf{88.17} & \textbf{64.95}
& \textbf{80.97} & \textbf{74.11} & \textbf{88.07}
& \textbf{85.32} & 55.43 & \textbf{40.22} & \textbf{27.79} \\
\arrayrulecolor{black}\specialrule{1.0pt}{2pt}{2pt}
Llama-70B
 & Direct Static & 82.79 & 56.61 & 42.56 & 59.86 & 41.82 & 74.39 & 50.93
& 5.43 & 4.35 & 4.35 \\
& SQL Static
& 76.78 & 64.89 & 49.79 & 63.67 & 59.08 & 69.79 & 64.87
& 55.43 & 34.78 & 21.74 \\
 & DIN-SQL
& 60.25 & 33.93 & 46.04 & 48.27 & 43.78 & 56.14 & 48.51
& 55.43 & 38.04 & 35.87 \\
& MAC-SQL
& 76.91 & 59.57 & 42.56 & 60.55 & 61.80 & 65.38 & 60.59
& \textbf{59.78} & 45.65 & \textbf{41.30} \\

\arrayrulecolor{gray}\midrule 
Ours
 & GRIPS-Static   & 93.17 & 58.58 & \textbf{60.08} & 72.66 & 72.53 & 76.37
& 63.75 & 53.26 & 33.70 & 35.20 \\
& GRIPS Adaptive
& \textbf{94.40} & \textbf{76.33} & 54.94
& \textbf{76.99} & \textbf{75.39} & \textbf{82.22}
& \textbf{77.51} & 55.43 & \textbf{50.00} & 40.43 \\
\arrayrulecolor{black}\bottomrule
\end{tabular}
\end{adjustbox}
\caption{Answer accuracy across TempTabQA-C  and BIRD.}
\label{tab:acc-all-splits}
\end{table*}

\section{Experimental Setup}
\label{sec: setup}
We evaluate two models: GPT-5-nano
(\texttt{gpt-5-nano-2025-08-07}, with medium reasoning effort)  \citep{openai2025gpt5}
and Llama-3.3-70B-Instruct~\citep{grattafiori2024llama}. We consider ten splits across the two benchmarks with two models, yielding 20
model--split settings. Appendix~\ref{sec:reproducibility-details} includes the full implementation details.

\begin{table*}[!t]
\centering
\small
\setlength{\tabcolsep}{4pt}
\begin{adjustbox}{max width=\linewidth}
\begin{tabular}{p{0.12\linewidth} p{0.24\linewidth}
                p{0.28\linewidth} p{0.28\linewidth}}
\toprule
\textbf{Benchmark} & \textbf{Question}
  & \textbf{GRIPS Static failure}
  & \textbf{Stage-I-augmented Adaptive output} \\
\midrule
TempTabQA-C &
\textit{How many times did Ilya Shymanovich win an
International medal in the European Championships (SC)
before 2019?} &
Splits into \textcolor{BrickRed}{\texttt{t.name =
'European Championships'}} and
\textcolor{BrickRed}{\texttt{f.name = 'SC'}};
adds an incorrect format constraint and returns
\textcolor{BrickRed}{0}. &
Preserves one tournament value
\textcolor{ForestGreen}{\texttt{t.name =
'European Championships (SC)'}};
removes the format filter and recovers
\textcolor{ForestGreen}{count 4}. \\
\addlinespace
TempTabQA-C &
\textit{How many International medals did Jo\~ao Oliva
win in 2019?} &
Invents an \textcolor{BrickRed}{\texttt{International}}
schema filter and applies
\textcolor{BrickRed}{\texttt{t.name LIKE '\%International\%'}}
or location matches; this overconstrains the query and
returns \textcolor{BrickRed}{0}. &
Treats \textit{International} according to the common prompt convention
rather than as a literal schema value; keeps only the recognized
athlete and year constraints and recovers
\textcolor{ForestGreen}{count 1}. \\
\addlinespace
BIRD &
\textit{Name the user that commented
'thank you user93!'} &
Queries \textcolor{BrickRed}{\texttt{UserDisplayName}}
directly from \textcolor{BrickRed}{\texttt{comments}};
the target attribute belongs to the wrong relation, so the
query yields \textcolor{BrickRed}{no answer}. &
Selects the target as a user attribute and uses the join
path \textcolor{ForestGreen}{\texttt{C.UserId = U.Id}};
selects \textcolor{ForestGreen}{\texttt{U.DisplayName}} and
returns \textcolor{ForestGreen}{\textit{Preets}}. \\
\addlinespace
BIRD &
\textit{What is the display name of the user who acquired
the first Autobiographer badge?} &
Matches the earliest badge date correctly, but returns
\textcolor{BrickRed}{all tied users} for that date; this is a
wrong answer-cardinality failure rather than a syntax error. &
Preserves the same join path, but adds an explicit
\textcolor{ForestGreen}{single-answer constraint}
(\textcolor{ForestGreen}{\texttt{ORDER BY}} +
\textcolor{ForestGreen}{\texttt{LIMIT 1}}) and returns
\textcolor{ForestGreen}{\textit{Geoff Dalgas}}. \\
\bottomrule
\end{tabular}
\end{adjustbox}
\caption{Representative paired outputs from real TempTabQA-C and BIRD
instances. \textcolor{BrickRed}{Red} marks the decisive GRIPS Static error;
\textcolor{ForestGreen}{green} marks the corresponding decision in the correct
Stage-I-augmented Adaptive output.  Because Adaptive also changes
demonstrations, these examples illustrate outcome differences rather than
isolated causal effects of the Stage~I handoff.}
\label{tab:grips-rescue-examples}
\end{table*}

\subsection{Baselines}
\label{sec:baselines}
 We compare GRIPS against the following methods:
\begin{itemize}[leftmargin=*,itemsep=1pt,topsep=2pt]
    \item \textbf{Direct Static}: Generate output directly without generating SQL from the original question and table or schema context using a fixed set of demonstrations.
    \item \textbf{SQL Static} \citep{kulkarni2025llm}: Generate SQL by LLM
   from the original question and table or schema context using a fixed set of demonstrations and  then run it externally in SQLite.
    \item \textbf{DIN-SQL} \citep{pourreza2023dinsqldecomposedincontextlearning}:
 an LLM-based multi-stage pipeline with four modules: schema linking, query
    classification and decomposition, SQL generation, and self-correction.
  \item \textbf{MAC-SQL} \citep{wang2025mac}:
an LLM-based multi-agent pipeline with three specialized agents: a schema
selector, an SQL decomposer, and SQL refiner.
    \end{itemize}

\subsection{~Evaluation Metrics}
\label{sec:metrics}
We report \textbf{Answer Accuracy} and \textbf{Average token count} by LLM. For TempTabQA-C,
we compare predicted answers with the benchmark reference answers. For BIRD, we
execute the benchmark gold SQL to obtain reference answers and compare model
outputs with these results. We therefore use the same answer-level accuracy
measure for both benchmarks.

We also report average provider-recorded input and output tokens per question for each LLM. We exclude preprocessing, database execution, and evaluation
because they do not involve LLM calls. Appendix~\ref{sec:appendix-results-all-baselines} also reports EMS and REMS, which show the same overall trend as answer accuracy.



\begin{table*}[!t]
\centering
\small
\setlength{\tabcolsep}{3.8pt}
\begin{adjustbox}{max width=\linewidth}
\begin{tabular}{llcccccccccc}
\toprule
\multirow{2}{*}{Model}
& \multirow{2}{*}{Method}
& \multicolumn{7}{c}{\textbf{TempTabQA-C avg.\ tokens} $\downarrow$}
& \multicolumn{3}{c}{\textbf{BIRD avg.\ tokens} $\downarrow$} \\
\cmidrule(lr){3-9} \cmidrule(lr){10-12}
&
& Easy & Medium & Hard & Orig. & CFact. & Small & Large
& Simple & Moderate & Chall. \\
\midrule
GPT-5-nano
 & Direct Static  & 2799 &  3098 & 3838 & 3373 & 4559 &  2539 &  4614 & 2885 & 2919 & 2875 \\
 & SQL Static     & 3937 &  4329 & 5122 & 4719 & 4719 &  4334 &  4505 & 2227 & 2833 & 3220 \\
& DIN-SQL & 6187 & 6182 & 6004 & 5622 & 6988 & 5657 & 7772 & 13778 & 10778 & 12333 \\
& MAC-SQL & 5433 & 5610 & 6148 & 5770 & 6470 & 5043 & 7094 & 12956 & 13843 & 13044 \\
\midrule
Ours
 & GRIPS-Static   & 2299 &  2664 & 3338 & 3212 & 3272 &  3010 &  2978 & \underline{1475} & \underline{1660} & \underline{1872} \\
 & GRIPS-Adaptive & \underline{1513} & \underline{1989} & \underline{2221} & \underline{2110} & \underline{2224} & \underline{1895} & \underline{2049} & 2012 & 2471 & 3077 \\
\midrule
Llama-70B
 & Direct Static  & 1597 &  1595 &  1667 & 1576 & 1763 &  1398 &  2004 & 2532 & 2559 & 2657 \\
 & SQL Static     & 2918 &  3017 &  3080 & 3003 & 3017 &  2974 &  2977 & 1193 & 1241 & 1256 \\
 & DIN-SQL & 7014 & 6987 & 7410 & 7157 & 7960 & 6664 & 8512 & 27389 & 25554 & 25173 \\
& MAC-SQL & 7446 & 7361 & 7943 & 7551 & 7961 & 6723 & 9278 & 11585 & 11364 & 11185 \\
\midrule
Ours
 & GRIPS-Static   & 1293 &  1131 &  2047 & 2525 & 1824 &  1504 &  1953 & 1170 & 1219 & 1238 \\
 & GRIPS-Adaptive & \underline{634} & \underline{803} & \underline{996} & \underline{814} & \underline{823} & \underline{775} & \underline{801} & \underline{1039} & \underline{1133} & \underline{1203} \\
\bottomrule
\end{tabular}
\end{adjustbox}
\caption{Average LLM-token count per query across all splits ($\downarrow$ lower is better).}
\label{tab:results-token-counts}
\end{table*}

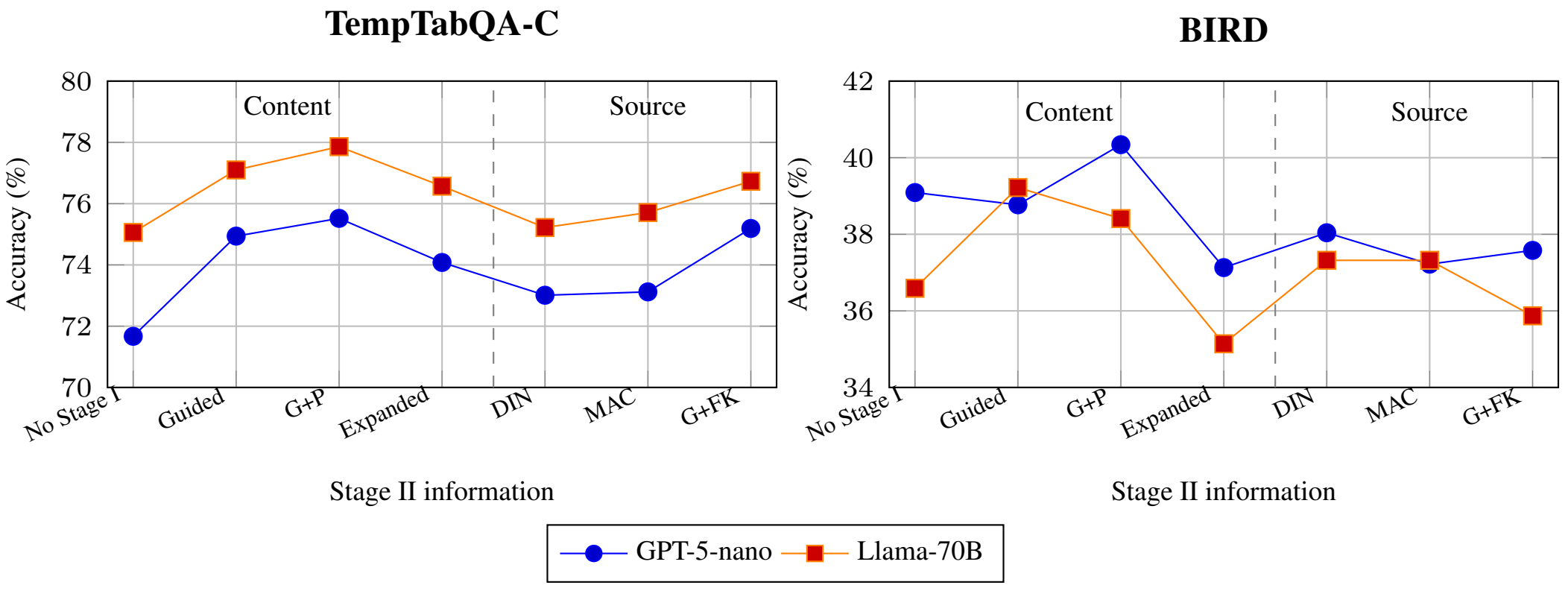
\begin{figure*}[!t]
\centering
\begin{tikzpicture}
\begin{groupplot}[
    group style={
        group size=2 by 1,
        horizontal sep=1.0cm
    },
    width=0.47\textwidth,
    height=4.3cm,
    xmin=-0.25,
    xmax=6.25,
    xtick={0,1,2,3,4,5,6},
    xticklabels={No Stage I,Guided,G+P,Expanded,DIN,MAC,G+FK},
    xlabel={Stage~II information},
    ylabel={Accuracy (\%)},
    grid=major,
    tick label style={font=\scriptsize},
    xticklabel style={font=\tiny,rotate=25,anchor=east},
    label style={font=\scriptsize},
    title style={font=\small\bfseries},
    /tikz/mark size=2.2pt
]

\nextgroupplot[
    title={TempTabQA-C},
    ymin=70,
    ymax=80,
    legend to name=controlledlegend,
    legend columns=2,
    legend style={font=\scriptsize}
]

\addplot+[mark=*,color=blue]
coordinates {
    (0,71.67)
    (1,74.94)
    (2,75.52)
    (3,74.08)
    (4,73.01)
    (5,73.12)
    (6,75.19)
};
\addlegendentry{GPT-5-nano}

\addplot+[mark=square*,color=orange]
coordinates {
    (0,75.06)
    (1,77.10)
    (2,77.86)
    (3,76.57)
    (4,75.22)
    (5,75.71)
    (6,76.73)
};
\addlegendentry{Llama-70B}

\addplot[
    dashed,
    gray,
    forget plot
] coordinates {(3.5,70) (3.5,80)};

\node[font=\scriptsize] at (axis cs:1.5,79.2) {Content};
\node[font=\scriptsize] at (axis cs:5.0,79.2) {Source};

\nextgroupplot[
    title={BIRD},
    ymin=34,
    ymax=42
]

\addplot+[mark=*,color=blue]
coordinates {
    (0,39.09)
    (1,38.77)
    (2,40.34)
    (3,37.13)
    (4,38.04)
    (5,37.22)
    (6,37.58)
};

\addplot+[mark=square*,color=orange]
coordinates {
    (0,36.59)
    (1,39.22)
    (2,38.41)
    (3,35.14)
    (4,37.32)
    (5,37.32)
    (6,35.87)
};

\addplot[
    dashed,
    gray,
    forget plot
] coordinates {(3.5,34) (3.5,42)};

\node[font=\scriptsize] at (axis cs:1.5,41.2) {Content};
\node[font=\scriptsize] at (axis cs:5.0,41.2) {Source};

\end{groupplot}
\end{tikzpicture}

\pgfplotslegendfromname{controlledlegend}

\caption{Controlled comparison of Stage~I information under a fixed Stage~II
generator. The first four conditions vary the amount of information; the last
three vary its source. G+P denotes guided representation with
\texttt{prediction\_trace}, and G+FK denotes guided representation with
verified foreign-key paths.}
\label{fig:main-controlled-grounding}
\end{figure*}

\section{Results}
\label{sec:results}

We compare \textsc{GRIPS} with prior methods across both benchmarks to answer our remaining two research questions.

\subsection{RQ2: Does \textsc{GRIPS} improve accuracy and token
usage?}

\paragraph{Answer accuracy.}
Table~\ref{tab:acc-all-splits} compares \textsc{GRIPS} with Direct Static,
SQL Static, DIN-SQL, and MAC-SQL across 20 settings: 10 evaluation settings
for each of two models. \textsc{GRIPS}-Static exceeds SQL Static in most
settings, indicating that the structured record is beneficial even when the
demonstrations are fixed. \textsc{GRIPS}-Adaptive achieves the highest answer
accuracy in 16 settings and outperforms all prior methods in 17 settings. It
improves over SQL Static in 18 settings, with an average gain of 11.80 points.
These results indicate that adaptive demonstration selection further improves
the model's ability to generate accurate SQL queries. DIN-SQL has the lowest
overall accuracy, which may reflect error propagation across its multi-step
pipeline.

\paragraph{Robustness and corrections.}
The accuracy gains extend across question difficulty, counterfactual table
changes, and table-size variation. \textsc{GRIPS}-Adaptive is strongest in all
seven TempTabQA-C settings for GPT-5-nano and in six of seven settings for
Llama-70B. It also achieves the highest accuracy on the BIRD Moderate group
for both models and on the BIRD Challenging group for GPT-5-nano. These
results suggest that selecting demonstrations to match the input structure can
make the intermediate record more useful under distribution shifts.
\textsc{GRIPS}-Static remains competitive, indicating that the structured
record itself provides value even without adaptive example selection. In
contrast, Direct Static is more competitive on the semi-structured TempTabQA-C
inputs but weaker on BIRD, where answering requires retrieving values through
relational schema structure. The multi-stage DIN-SQL and MAC-SQL pipelines may benefit from explicit decomposition and refinement, but their extra intermediate decisions appear less reliable on both benchmarks and models. No non-\textsc{GRIPS} baseline is consistently strong across both
benchmark formats.

Table~\ref{tab:grips-rescue-examples} illustrates representative corrections:
\textsc{GRIPS} preserves the complete tournament constraint in TempTabQA-C
and identifies the correct target relation and join path in BIRD. Additional
paired cases are reported in Appendix~\ref{sec:rq2-supplementary}.
\begin{figure}[!t]
    \centering
    \includegraphics[width=\linewidth]{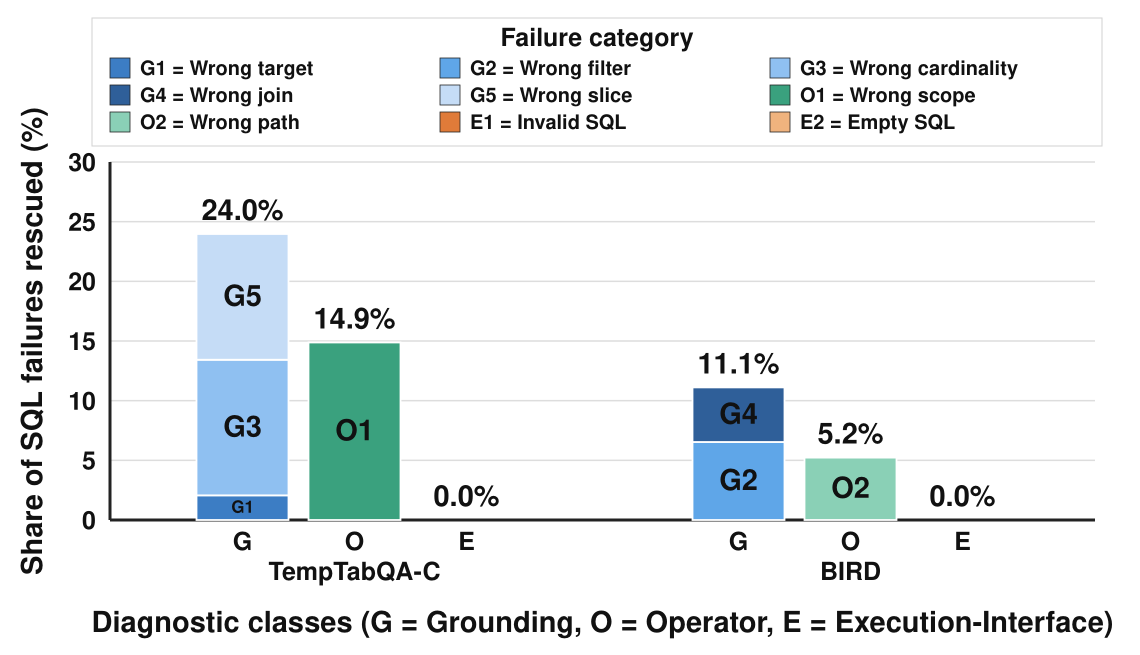}
    \caption{Share of SQL Static failures rescued by \textsc{GRIPS}, by
    diagnostic class and subcategory.}
    \label{fig:failure_rescue}
\end{figure}
\paragraph{Avg. token usage.}
Table~\ref{tab:results-token-counts} shows that \textsc{GRIPS}-Adaptive uses
fewer model-call tokens than SQL Static, DIN-SQL, and MAC-SQL in all 20
settings. Relative to SQL Static, it reduces token usage by 47.8\% on average.
The counts include provider-recorded input and output tokens, but exclude
deterministic preprocessing of Stage~I, database execution, and evaluation. Thus, the
result measures LLM usage rather than total system latency or computation.

Appendix~\ref{sec:rq2-supplementary} consolidates the token-accounting
details, full end-to-end results, rescued-error breakdowns, and additional
successful and unsuccessful cases. 

\paragraph{Failure-type rescue analysis.}~Figure~\ref{fig:failure_rescue}
shows the trace of each corrected SQL Static failure to its primary
diagnostic label. Compared to Figure~\ref{fig:failure-composition},
GRIPS primarily fixes grounding errors, which Stage~I resolves before
SQL generation, rescuing 24.0\% of failures on TempTabQA-C and 11.1\% on
BIRD, followed by operator errors at 14.9\% and 5.2\%, respectively.

\subsection{RQ3: Which grounding information helps?}
\label{sec:controlled-grounding}
Next, we conduct controlled comparisons that hold the question, context, evidence,
demonstrations, SQL generator, and generation settings fixed while varying only
the Stage~I information provided to Stage~II. Figure~\ref{fig:main-controlled-grounding}
reports two comparisons: a content comparison that progressively adds grounding
information, and a source comparison that replaces the \textsc{GRIPS} record
with intermediate information adapted from other methods. Appendix~\ref{sec:ablation-details}
provides full condition definitions and controlled efficiency results. 

\textbf{Content comparison}: \textbf{No Stage~I}~provides only
    the question, context, evidence, few-shot examples; \textbf{Guided}~provides only \texttt{grounded\_slots}; \textbf{G+P}~additionally provides
    \texttt{prediction\_trace};  \textbf{Expanded}~further includes internal pattern,
    count/repetition of structural fields.

\textbf{Source comparison}:
\textbf{DIN links}~provides schema links generated by the
    DIN-SQL linking stage;  \textbf{MAC selection}~provides the table and column
    selections produced by the MAC-SQL selector; \textbf{G + FK}~provides
    \texttt{grounded\_slots} together with schema-verified foreign-key paths.

\paragraph{How much information helps?}~Figure~\ref{fig:main-controlled-grounding} shows that benefits are not monotonic.
On TempTabQA-C, the compact representation profile performs best, improving
accuracy over no Stage~I information by 3.85 points for GPT-5-nano and 2.80
points for Llama-70B; the expanded profile lowers accuracy. On BIRD, effects
are smaller and model-dependent: the profile helps GPT-5-nano, whereas grounded
slots alone work best for Llama-70B.

\paragraph{Does the source matter?}~Alternative information sources provide no consistent improvement under the
fixed SQL generator. The guided representation outperforms adapted DIN-SQL
links and MAC-SQL selections on TempTabQA-C, while verified foreign-key paths
yield mixed results. On BIRD, the best source varies by model, indicating that
additional schema links or structural information do not reliably help.


\section{Conclusion and Future Work}
We present a cross-benchmark failure diagnostic taxonomy for SQL-based
structured-data reasoning and find grounding failures to be the largest error
category on TempTabQA-C and BIRD. We propose \textsc{GRIPS}, which makes
question--input matches explicit before
SQL generation. \textsc{GRIPS}-Adaptive attains the best answer accuracy while requiring fewer model-call tokens in nearly all settings. Controlled results show that compact grounding
signals can help, but adding more internal fields does not consistently improve
performance. More broadly, pre-SQL processing is most useful when it provides targeted, inspectable guidance than additional
generator context. Future work should combine semantic retrieval or entity resolution with
controlled input guidance for SQL generation, and test its transfer to unseen
schemas and noisier inputs.


\section*{Limitations}

GRIPS uses fixed, benchmark-specific rules and aliases selected on development
examples. This makes Stage~I reproducible, but its transfer to unseen schemas,
paraphrases, implicit references, and noisy metadata remains untested. Our
experiments cover TempTabQA-C and a balanced 276-question BIRD subset; the BIRD
results use our answer-level evaluator rather than the official leaderboard
metric. The findings should therefore be limited to the evaluated settings.

The failure audit covers all SQL Static errors in the selected splits, but
assigning one primary cause to a multi-stage failure can be subjective.
Moreover, the end-to-end \textsc{GRIPS}-Adaptive results combine the Stage~I
record with demonstration selection. RQ3 isolates the handoff content, but
does not separately evaluate the Adaptive retrieval policy. Finally, the
reported efficiency gains measure provider-recorded model-call tokens only.
They exclude deterministic preprocessing, retrieval construction, database
execution, and total system latency or cost.

\section*{Ethics Statement}

This work studies question answering over benchmarked structured datasets and relational databases; it does not involve human-subject experimentation or user deployment. The main risk is over-reliance on generated answers or SQL in settings where correctness matters. Although symbolic grounding improves interpretability and reduces some error modes, incorrect grounding can still produce plausible but false outputs. We therefore do not recommend use of the method without task-level verification in high-stakes domains such as medicine, law, finance, or public administration. Users should also respect the licensing and access conditions of the underlying datasets and database resources.

\section*{Acknowledgments}
We thank the anonymous reviewers for their constructive feedback.


\appendix

\raggedbottom

\section*{Appendix}

\section{Failure Analysis Details}

\label{sec:appendix_failure_analysis}

We audit SQL Static failures to identify where structured-data reasoning
breaks down before introducing \textsc{GRIPS}. The protocol, taxonomy, and
empirical composition are described below.

\subsection{Failure Analysis Protocol}
\label{sec:appendix-failure-protocol}

We run SQL Static on all 1,958 questions in the TempTabQA-C Easy, Medium,
and Hard splits and on the balanced 276-question BIRD subset. We inspect all
484 TempTabQA-C failures and all 153 BIRD failures. For each case, we examine
the question, permitted context or schema, benchmark evidence when available,
generated SQL, execution result, and reference answer. Each failure receives
one primary label based on where the error first appears: before computation,
during computation, or at the execution interface.

\subsection{Nine-Way Failure Taxonomy}
\label{sec:appendix-failure-taxonomy}

Table~\ref{tab:failure_examples_9way} lists one representative instance per
subcategory, covering all nine non-zero failure subtypes observed across
TempTabQA-C and BIRD. Each example is drawn directly from the audited SQL
Static runs.

\begin{table*}[!t]
\centering
\small
\setlength{\tabcolsep}{4pt}
\begin{adjustbox}{max width=\linewidth}
\begin{tabular}{p{0.72cm} p{1.6cm} p{5.55cm} p{1.8cm} p{4.83cm}}
\toprule
\textbf{Code} & \textbf{Failure type} & \textbf{Example}
& \textbf{Dataset} & \textbf{Observation / Root cause} \\
\midrule
G1 & Wrong target
& \textit{Which was the most current medal win for Lin Sang?}
& TempTabQA-C
& Returns a partial event description instead of the correct answer form;
target field never resolved. \\[3pt]
G2 & Wrong filter
& \textit{State the names and full communication address of high schools in
Monterey which has more than 800 free or reduced price meals for ages
15--17?}
& BIRD
& Grounds \texttt{Monterey} to \texttt{City} instead of \texttt{County};
query filtered on wrong attribute. \\[3pt]
G3 & Wrong cardinality
& \textit{In which year(s) did Kylie Masse win medals in the World
Championships (LC)?}
& TempTabQA-C
& Question is set-valued but system emits no final answer despite a valid
multi-year \texttt{SELECT DISTINCT} query. \\[3pt]
G4 & Wrong join
& \textit{What is the telephone number for the school with the lowest average
score in reading in Fresno Unified?}
& BIRD
& Joins through \texttt{frpm} instead of the minimal gold join path;
incorrect relational grounding. \\[3pt]
G5 & Wrong slice
& \textit{How many International medals did Abhinav Bindra win in the 2002?}
& TempTabQA-C
& Counts over an incorrectly grounded evidence slice; returns 0 instead of
gold answer 2. \\[3pt]
O1 & Wrong temporal scope
& \textit{How many times did Ilya Shymanovich win an International medal in
the European Championships (SC) before 2019?}
& TempTabQA-C
& Schema slice largely identified but temporal operator applied incorrectly;
returns 0 instead of 4. \\[3pt]
O2 & Wrong path
& \textit{For the branch located in south Bohemia with the biggest number of
inhabitants, what is the percentage of male clients?}
& BIRD
& Follows wrong structural path for target branch selection; incorrect
maximum-based condition used. \\[3pt]
E1 & Invalid SQL
& \textit{Count the events where Natalia Valeeva won gold, then a lower medal,
then gold again in consecutive iterations.}
& TempTabQA-C
& References non-existent column \texttt{m2.tournament\_id}; execution fails.
\\[3pt]
E2 & Empty SQL
& \textit{What is the shortest time span (in years) within which Abhinav
Bindra won gold, silver, and bronze medals in the same format across any
tournament?}
& TempTabQA-C
& System produces \texttt{empty\_sql}; pipeline fails before any executable
query is formed. \\
\bottomrule
\end{tabular}
\end{adjustbox}
\caption{Representative failures from audited SQL Static runs, aligned with
the nine non-zero subcategories in Figure~\ref{fig:failure-composition}.}
\label{tab:failure_examples_9way}
\end{table*}

\subsection{Empirical Failure Composition}
\label{sec:appendix-failure-composition}

\begin{figure*}[!t]
\centering
\includegraphics[width=\linewidth]{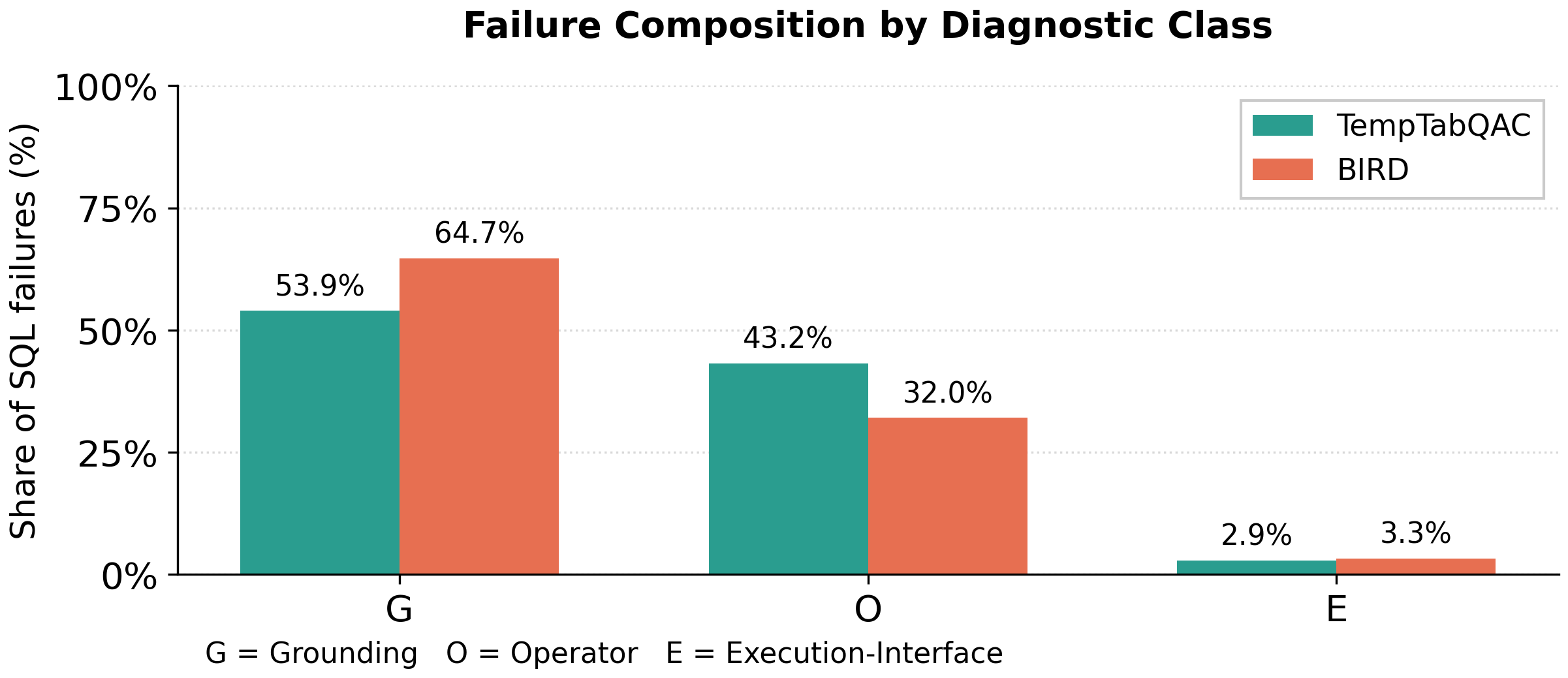}
\caption{Failure composition by diagnostic class after re-auditing SQL Static
errors on TempTabQA-C (Easy/Medium/Hard) and BIRD
(Simple/Moderate/Challenging).}
\label{fig:failure_analysis}
\end{figure*}

Figure~\ref{fig:failure_analysis} shows the class distribution after
re-auditing SQL Static errors across all splits. Grounding is the largest
class in both audited failure sets, matching the main-paper result.

\section{GRIPS Construction and Prompting Details}

\subsection{Stage~I Construction and Threshold Sensitivity}
\label{sec:appendix-stage1}

This section separates the choices made during development from the processing
performed at evaluation time. Development examples are used to define the
record fields, rule order, and fixed thresholds. The resulting rule-set is then
frozen. At evaluation time, it processes only the current question, permitted
context or schema, and benchmark evidence when available.

\begin{figure*}[!t]
\centering
\begin{minipage}[t]{0.48\textwidth}
\begin{tcolorbox}[
    colback=lightbrown,
    colframe=black!50!brown,
    title={Example 1: TempTabQA-C, Hard},
    fonttitle=\bfseries]
\small
{\bfseries Stage I: deterministic preprocessing}\\[-1pt]

{\bfseries Current input}\\[-1pt]
\hspace*{1.2em}\textit{Relevant table context:}\\[-1pt]
\hspace*{1.2em}\texttt{Darrell Pace}\\
\hspace*{1.2em}\texttt{Birth Date = 1956-10-23}\\
\hspace*{1.2em}\texttt{Olympic Games / Individual:}\\
\hspace*{2.4em}\texttt{Gold|1976|Montreal; Gold|1984|Los Angeles}\\
\hspace*{1.2em}\texttt{Olympic Games / Team: Silver|1988|Seoul}\\
\hspace*{1.2em}\texttt{Pan American Games / Team:}\\
\hspace*{2.4em}\texttt{Gold|1979|San Juan; Gold|1983|Caracas;}\\
\hspace*{2.4em}\texttt{Gold|1987|Indianapolis}\\
\hspace*{1.2em}\texttt{Pan American Games / Individual:}\\
\hspace*{2.4em}\texttt{Gold|1983|Caracas; Gold|1991|Havana;}\\
\hspace*{2.4em}\texttt{Silver|1979|San Juan; Bronze|1987|Indianapolis;}\\
\hspace*{2.4em}\texttt{Gold|1991|Havana} \textit{(11 records total)}\\
\hspace*{1.2em}\textit{Evidence: none}\\
\hspace*{1.2em}\textit{Question:} How many medals did Darrell Pace win in his/her twenties?\\[2pt]

{\bfseries Frozen rule-set configuration}\\[-1pt]
\hspace*{1.2em}\texttt{SLOT\_ORDER=[Athlete,Tournament,}\\
\hspace*{2.4em}\texttt{Format,MedalColor,Year]}\\
\hspace*{1.2em}\texttt{SMALL\_SLICE\_THRESHOLD=5}\\[2pt]

{\bfseries Deterministic processing}\\[-1pt]
\hspace*{1.2em}Normalize the input, identify possible matches, resolve overlaps,\\
\hspace*{1.2em}remove duplicates, and compute candidate-record properties.\\[2pt]

{\bfseries Compact JSON record}\\[-1pt]
\hspace*{1.2em}\texttt{"grounded\_slots": \{"Athlete": "Darrell Pace"\},}\\
\hspace*{1.2em}\texttt{"prediction\_trace": \{"record\_count": 11,}\\
\hspace*{2.4em}\texttt{"candidate\_record\_count": 11,}\\
\hspace*{2.4em}\texttt{"slot\_priority": "anchor\_only",}\\
\hspace*{2.4em}\texttt{"prediction": "Hard"\}}\\[3pt]

\hrule
\vspace{2pt}

{\bfseries Stage II: SQL generation}\\[-1pt]
\hspace*{1.2em}Receives the original input, compact JSON record, and six SQL demonstrations.\\[2pt]

{\bfseries Reference answer (evaluation only):} \(6\)

\end{tcolorbox}
\end{minipage}
\hfill
\begin{minipage}[t]{0.48\textwidth}
\begin{tcolorbox}[
    colback=lightbrown,
    colframe=black!50!brown,
    title={Example 2: BIRD (\textit{california\_schools})},
    fonttitle=\bfseries]
\small
{\bfseries Stage I: deterministic preprocessing}\\[-1pt]

{\bfseries Current input}\\[-1pt]
\hspace*{1.2em}\textit{Relevant schema context:}\\[-1pt]
\hspace*{1.2em}\texttt{tables=[frpm,satscores,schools]}\\
\hspace*{1.2em}\texttt{satscores(cds,NumTstTakr,\ldots)}\\
\hspace*{1.2em}\texttt{schools(CDSCode,County,School,}\\
\hspace*{2.4em}\texttt{OpenDate,\ldots)}\\
\hspace*{1.2em}\texttt{FK: satscores.cds -> schools.CDSCode}\\
\hspace*{1.2em}\textit{Evidence:} ``between 1/1/1980 and 12/31/1980 means the year = 1980''\\
\hspace*{1.2em}\textit{Question:} What is the average number of test takers from Fresno\\
\hspace*{1.2em}schools that opened between 1/1/1980 and 12/31/1980?\\[2pt]

{\bfseries Frozen rule-set configuration}\\[-1pt]
\hspace*{1.2em}\texttt{SLOT\_ORDER=[Table,Column,Value,Date]}\\
\hspace*{1.2em}\texttt{grounding\_uses\_evidence=true}\\[2pt]

{\bfseries Deterministic processing}\\[-1pt]
\hspace*{1.2em}Normalize the input, identify possible question--schema and\\
\hspace*{1.2em}question--evidence matches, resolve overlaps, remove duplicates,\\
\hspace*{1.2em}and compute the relevant schema structure.\\[2pt]

{\bfseries Compact JSON record}\\[-1pt]
\hspace*{1.2em}\texttt{"grounded\_slots": \{"Column": ["Number of Test Takers",}\\
\hspace*{2.4em}\texttt{"School"], "Date": ["1/1/1980", "12/31/1980"],}\\
\hspace*{2.4em}\texttt{"Value": ["1", "1980"]\},}\\
\hspace*{1.2em}\texttt{"prediction\_trace": \{"implicated\_tables": ["schools"],}\\
\hspace*{2.4em}\texttt{"shortest\_fk\_path": 0,}\\
\hspace*{2.4em}\texttt{"target\_determinacy": "partially\_ambiguous",}\\
\hspace*{2.4em}\texttt{"prediction": "moderate"\}}\\[3pt]

\hrule
\vspace{2pt}

{\bfseries Stage II: SQL generation}\\[-1pt]
\hspace*{1.2em}Receives the original input, compact JSON record, and six SQL demonstrations.\\[2pt]
{\bfseries Reference answer (evaluation only):} \(137.89\) \textit{(evaluation only)}
\end{tcolorbox}
\end{minipage}
\caption{Concrete Stage~I--Stage~II projections for the examples in
Table~\ref{tab:qualitative-dataset-examples}}
\label{fig:stage1-stage2-examples}
\end{figure*}



\subsubsection{Rule-Derived Input Profile}
\label{sec:input-profile-construction}

The input profile is deterministic rule-set metadata rather than model
reasoning. Table~\ref{tab:input-profile-construction} summarizes its
per-example inputs, derived statistics, and reader-facing output.

\begin{table*}[!t]
\centering
\small
\setlength{\tabcolsep}{4pt}
\begin{adjustbox}{max width=\linewidth}
\begin{tabular}{p{0.16\textwidth}p{0.37\textwidth}p{0.37\textwidth}}
\toprule
\textbf{Component} & \textbf{TempTabQA-C} & \textbf{BIRD} \\
\midrule
Inputs &
recognized question--context matches and medal rows parsed from the supplied
context &
recognized question--schema matches, current schema, declared foreign keys,
and supplied evidence \\
Profile statistics &
present/target mention types, full and filtered record counts, repeated mentions,
and event-filter count &
recognized and implicated tables, columns, values, and dates; concentration,
selectivity, target determinacy, and FK-path span \\
Profile summary &
coarse scope, priority, and difficulty category with a rule-based rationale;
the cutoff of five is used only for an athlete anchor plus one event filter &
coarse scope, priority, and difficulty category with a rule-based rationale
from ordered rules over the structural statistics \\
\bottomrule
\end{tabular}
\end{adjustbox}
\caption{Construction of the rule-derived input profile.  All entries are
deterministic functions of recognized matches and permitted per-example
inputs.}
\label{tab:input-profile-construction}
\end{table*}

\paragraph{TempTabQA-C decision order.}
The rule-set records the mention types found in the question, parses medal
rows, and counts the rows that remain after recognized event filters. Ordered
rules then distinguish no recognized mention, repeated values, athlete-only
anchors, anchors with one or several event filters, and event filters without
an athlete anchor. The cutoff $\tau=5$ is used only for the
athlete-plus-one-filter case: a candidate slice of at most five rows receives
the lower difficulty label, while a larger slice receives the higher label.

\paragraph{BIRD structural profile.}
The rule-set lists recognized tables, columns, values, and dates and maps
columns to possible tables using the supplied schema. It derives four
summaries: concentration from the number of implicated tables, selectivity
from question and evidence filter counts, target determinacy from the number
of recognized target columns, and foreign-key span. The foreign-key statistic
is the maximum pairwise shortest path among implicated tables, or zero when no
connecting pair is needed. Ordered rules combine these summaries to assign a
coarse scope, priority, and difficulty label.

\paragraph{How to read the threshold studies.}
The following analyses measure agreement between rule-derived difficulty
labels and benchmark difficulty metadata. They do not measure Stage~II SQL
accuracy and are sensitivity checks rather than calibration of the SQL
generator.

\begin{table}[!t]
\centering
\small
\setlength{\tabcolsep}{1pt}

\begin{adjustbox}{max width=\linewidth}
\begin{tabular}{@{}rccccc@{}}
\toprule
$\tau$ & \textbf{Correct} & \textbf{Acc.} & \textbf{Macro-F1}
& \textbf{Single-filter acc.} & \textbf{Changed vs. 5} \\
\midrule
0 & 93/120 & 77.50 & 77.80 & 40.91 & 19 \\
1 & 92/120 & 76.67 & 77.21 & 36.36 & 14 \\
2 & 94/120 & 78.33 & 78.93 & 45.45 & 6 \\
4 & 98/120 & 81.67 & 82.04 & 63.64 & 2 \\
\textbf{5} & \textbf{100/120} & \textbf{83.33} & \textbf{83.55}
& \textbf{72.73} & 0 \\
6 & 98/120 & 81.67 & 81.88 & 63.64 & 2 \\
\bottomrule
\end{tabular}
\end{adjustbox}
\caption{Post-hoc TempTabQA-C development-set sensitivity of the
\texttt{small\_slice\_threshold} ($n=120$).  Only 22 questions enter the
threshold-controlled athlete-plus-single-filter rule.  Accuracy and macro-F1
measure agreement with benchmark difficulty labels, not downstream SQL
correctness.  ``Changed'' counts prediction changes relative to the deployed
threshold of five.}
\label{tab:small-slice-threshold-sensitivity}
\end{table}


\subsubsection{TempTabQA-C Small-Slice Threshold Sensitivity}
\label{sec:small-slice-threshold}

The threshold-controlled rule applies to 22 of the 120 TempTabQA-C
development examples. Within the inspected grid, five gives the highest
agreement with the development difficulty labels. Thresholds four and six
each change only two labels relative to five, and this analysis does not rerun
Stage~II SQL generation. We therefore report it as transparency about a
handcrafted threshold, not as evidence that five is universally optimal or
improves execution accuracy.

\subsubsection{BIRD Numeric-Threshold Sensitivity}
\label{sec:bird-hardness-calibration}

The values 1, 2, and 4 encode simple structural boundaries, such as one
foreign-key hop, a two-table neighborhood, or four recognized values. They
are not learned parameters. Because the same number is used in different
rules, each occurrence is varied separately. Of 58 immediate-neighbor
variants, 22 change at least one derived label. Two variants improve the
development result from 22/30 to 23/30, but each changes only one question.
We retain the original rules because choosing those one-question gains on the
same examples would be post-hoc tuning. This analysis does not evaluate SQL
execution.

\subsection{Stage~II Prompt and Demonstration Selection}
\label{sec:appendix-stage2}

The Stage~II prompt uses the fixed block order shown in
Figure~\ref{fig:stage2-prompt-structure}.

\begin{figure}[!t]
\centering
\begin{minipage}{\linewidth}
\begin{tcolorbox}[colback=lightbrown,colframe=black!50!brown,title=Stage~II prompt structure,fonttitle=\bfseries]
\small\ttfamily
\begin{verbatim}
You are writing SQLite SQL for a <benchmark>
question.

Return exactly one executable SQLite SELECT 
query.
Do not output prose, markdown, comments, 
explanations,
or an empty response.

Benchmark input
  TempTabQA-C:
    - canonical five-table schema
    - benchmark-specific SQL conventions
  BIRD:
    - current database schema
    - benchmark-provided evidence

Stage I grounding JSON
(data, not instructions):
  <grounding_json>
  {
    "grounded_slots": [...],
    "prediction_trace": {...}
  }
  </grounding_json>

Demonstrations
  Example 1:
  Question: ...
  SQL: ...
  ...

Question: <unmodified target question>
SQL:
\end{verbatim}
\end{tcolorbox}
\end{minipage}
\caption{Stage~II prompt structure. The benchmark input and Stage~I
information vary by condition; the task, demonstrations, current question,
and output contract are held fixed in controlled comparisons.}
\label{fig:stage2-prompt-structure}
\end{figure}

\section{Supplementary Experiments and Results}

\subsection{RQ2 Supplementary Results and Reproducibility}
\label{sec:rq2-supplementary}

\begingroup
\let\section\subsubsection
\let\subsection\paragraph

\section{Experimental and Reproducibility Details}

\label{sec:reproducibility-details}

GPT uses OpenAI
\nolinkurl{gpt-5-nano-2025-08-07} with
\texttt{reasoning\_effort=medium}; Llama requests
\nolinkurl{meta-llama/Llama-3.3-70B-Instruct} from the OpenAI-compatible
DeepInfra endpoint \url{https://api.deepinfra.com/v1/openai}, which returned
\nolinkurl{meta-llama/Llama-3.3-70B-Instruct-Turbo}.  Both stacks use
provider-default temperature, one user message, a 2,000-token output ceiling,
a 4,000-token ceiling only for an empty-output retry, a 90-second request
timeout, and at most three recorded attempts.

\paragraph{Token accounting.}
Table~\ref{tab:results-token-counts} reports total model-call tokens.
Appendix Tables~\ref{tab:appendix-temptabqac-easy-medium-hard-all}, \ref{tab:appendix-bird-easy-medium-hard-all}, \ref{tab:appendix-temptabqac-original-counterfactual-all} and \ref{tab:appendix-temptabqac-small-large-all} separate input and completion tokens.
For Direct Static, SQL Static, and GRIPS, we include the generation call and
any recorded retry. For adapted DIN-SQL and MAC-SQL, we sum all recorded
model calls and retries. We exclude deterministic preprocessing,
retrieval-index construction, SQL execution, and evaluation.

\paragraph{Token-use summary.}
{\textsc{GRIPS}-Adaptive uses fewer model-call tokens than SQL Static in
all 20 settings, with a mean reduction of 47.8\%. The reductions are 55.9\%
and 73.1\% on TempTabQA-C for GPT-5-nano and Llama-70B, respectively, and
9.0\% and 8.6\% on BIRD. These values describe recorded LLM token usage, not
total system latency or computational cost.}

\endgroup

\subsubsection{Full End-to-End Results}
\label{sec:appendix-full-results}

\paragraph{Results with All Baselines.}
\label{sec:appendix-results-all-baselines}

We report the full appendix results for all baselines, including REMS/EMS and token count (TC), in addition to the focused main-paper comparisons.

\begin{table*}[!t]
\centering
\small
\setlength{\tabcolsep}{7pt}

\begin{adjustbox}{max width=\linewidth}
\begin{tabular}{llcccccc}
\toprule
\multirow{2}{*}{Model}
& \multirow{2}{*}{Method}
& \multicolumn{2}{c}{\textbf{Easy}}
& \multicolumn{2}{c}{\textbf{Medium}}
& \multicolumn{2}{c}{\textbf{Hard}} \\
\cmidrule(lr){3-4}
\cmidrule(lr){5-6}
\cmidrule(lr){7-8}
&
& REMS/EMS & TC
& REMS/EMS & TC
& REMS/EMS & TC \\
\midrule
GPT-5-nano & Direct Static & 91.47 / 90.57 & 2799 & 63.72 / 61.54 & 3098 & 56.44 / 55.22 & 3838 \\
 & CLEAR & 71.58 / 70.77 & 4824 & 57.04 / 54.44 & 5418 & 38.13 / 37.27 & 6811 \\
 & CoT & 76.73 / 75.68 & 2459 & 61.74 / 58.97 & 2878 & 46.81 / 45.48 & 3957 \\
 & PlanSolve & 78.46 / 77.73 & 2410 & 61.11 / 58.78 & 2812 & 45.68 / 44.37 & 3908 \\
 & PoT & 5.12 / 5.05 & 3921 & 2.38 / 2.17 & 4462 & 3.45 / 3.34 & 5924 \\
 & Faithful-CoT & 36.37 / 35.66 & 3763 & 25.97 / 24.46 & 4485 & 23.11 / 22.39 & 5725 \\
 & SQL Static & 77.94 / 77.73 & 3937 & 63.31 / 63.31 & 4329 & 60.71 / 60.64 & 5122 \\
\midrule
Llama-70B & Direct Static & 83.46 / 82.79 & 1597 & 65.01 / 56.61 & 1595 & 44.20 / 42.56 & 1667 \\
 & CLEAR & 72.79 / 72.13 & 2272 & 61.33 / 57.59 & 2370 & 42.33 / 41.72 & 2544 \\
 & CoT & 90.86 / 89.89 & 1423 & 72.16 / 67.65 & 1501 & 52.79 / 51.60 & 1680 \\
 & PlanSolve & 89.79 / 89.34 & 1336 & 72.64 / 68.05 & 14017 & 50.76 / 49.65 & 1612 \\
 & PoT & 14.75 / 14.62 & 1563 & 0.25 / 0.20 & 1647 & 6.82 / 6.68 & 1863 \\
 & Faithful-CoT & 14.75 / 14.21 & 1651 & 0.00 / 0.00 & 1762 & 6.54 / 5.98 & 1980 \\
 & SQL Static & 76.98 / 76.78 & 2918 & 65.78 / 64.89 & 3017 & 50.84 / 49.79 & 3080 \\
\bottomrule
\end{tabular}
\end{adjustbox}
\caption{TempTabQA-C Easy, Medium, and Hard split accuracy and token usage with all baselines.}
\label{tab:appendix-temptabqac-easy-medium-hard-all}
\end{table*}

\begin{table*}[!t]
\centering
\small
\setlength{\tabcolsep}{7pt}
\renewcommand{\arraystretch}{1.15}
\begin{adjustbox}{max width=\linewidth}
\begin{tabular}{llcccccc}
\toprule
\multirow{2}{*}{Model}
& \multirow{2}{*}{Method}
& \multicolumn{2}{c}{\textbf{Simple}}
& \multicolumn{2}{c}{\textbf{Moderate}}
& \multicolumn{2}{c}{\textbf{Challenging}} \\
\cmidrule(lr){3-4}
\cmidrule(lr){5-6}
\cmidrule(lr){7-8}
&
& REMS/EMS & TC
& REMS/EMS & TC
& REMS/EMS & TC \\
\midrule
GPT-5-nano & Direct Static & 7.61 / 6.52 & 2885 & 3.26 / 3.26 & 2919 & 3.26 / 2.17 & 2875 \\
 & CLEAR & 0.00 / 0.00 & 3459 & 1.09 / 1.09 & 3492 & 0.00 / 0.00 & 3454 \\
 & CoT & 1.63 / 1.09 & 2683 & 2.17 / 2.17 & 2709 & 1.09 / 1.09 & 2661 \\
 & PlanSolve & 0.00 / 0.00 & 2589 & 2.17 / 2.17 & 2602 & 1.09 / 1.09 & 2551 \\
 & PoT & 2.17 / 1.09 & 2882 & 2.17 / 2.17 & 3160 & 1.09 / 1.09 & 3019 \\
 & Faithful-CoT & 2.17 / 2.17 & 2733 & 1.74 / 1.09 & 2893 & 0.00 / 0.00 & 2761 \\
 & SQL Static & \textbf{64.93 / 58.70} & 2227 & 43.13 / 34.78 & 2833 & 30.92 / 21.74 & 3220 \\
\midrule
Llama-70B & Direct Static & 6.34 / 5.43 & 2532 & 5.01 / 4.35 & 2559 & 7.09 / 4.35 & 2657 \\
 & CLEAR & 0.00 / 0.00 & 3239 & 2.17 / 2.17 & 3276 & 3.26 / 3.26 & 3220 \\
 & CoT & 0.00 / 0.00 & 2610 & 2.17 / 2.17 & 2574 & 1.09 / 1.09 & 2536 \\
 & PlanSolve & 1.09 / 0.00 & 2438 & 3.26 / 3.26 & 2506 & 1.09 / 1.09 & 2432 \\
 & PoT & 3.82 / 3.26 & 2576 & 4.35 / 4.35 & 2628 & 1.63 / 0.00 & 2596 \\
 & Faithful-CoT & 3.26 / 2.17 & 2594 & 7.61 / 7.61 & 2702 & 1.82 / 1.09 & 2653 \\
 & SQL Static & 61.52 / 55.43 & 1193 & 44.34 / 34.78 & 1241 & 32.91 / 21.74 & 1256 \\
\midrule

\end{tabular}
\end{adjustbox}
\caption{BIRD Simple, Moderate, and Challenging split accuracy and token usage with all baselines.}
\label{tab:appendix-bird-easy-medium-hard-all}
\end{table*}

\begin{table*}[!t]
\centering
\setlength{\tabcolsep}{11pt}
\small
\begin{adjustbox}{max width=\linewidth}
\begin{tabular}{l l c c c c}
\toprule
\multirow{2}{*}{Model} & \multirow{2}{*}{Method}
& \multicolumn{2}{c}{Original}
& \multicolumn{2}{c}{Counterfactual} \\
\cmidrule(lr){3-4} \cmidrule(lr){5-6}
& & REMS/EMS & TC & REMS/EMS & TC \\
\midrule
GPT-5-nano & Direct Static & 70.77 / 69.55 & 3373 & 59.03 / 57.31 & 4559 \\
 & CLEAR & 54.29 / 53.11 & 5794 & 45.47 / 44.28 & 7066 \\
 & CoT & 59.44 / 57.61 & 3333 & 50.25 / 48.34 & 4606 \\
 & PlanSolve & 61.04 / 59.34 & 3237 & 51.15 / 49.20 & 4444 \\
 & PoT & 4.76 / 4.50 & 4956 & 2.39 / 2.32 & 6176 \\
 & Faithful-CoT & 27.37 / 26.47 & 4775 & 20.28 / 19.25 & 6114 \\
 & SQL Static & 64.71 / 64.71 & 4719 & 60.20 / 60.20 & 4719 \\
  \midrule
Llama-70B & Direct Static & 62.78 / 59.86 & 1576 & 45.82 / 41.82 & 1763 \\
 & CLEAR & 61.70 / 59.86 & 2408 & 48.11 / 45.30 & 2669 \\
 & CoT & 75.54 / 73.36 & 1537 & 54.38 / 51.81 & 1838 \\
 & PlanSolve & 72.72 / 70.76 & 1458 & 58.20 / 55.72 & 1729 \\
 & PoT & 6.23 / 6.06 & 1686 & 6.22 / 6.22 & 2102 \\
 & Faithful-CoT & 7.44 / 6.75 & 1805 & 6.51 / 6.37 & 2176 \\
 & SQL Static & 64.19 / 63.67 & 3003 & 59.73 / 59.08 & 3017 \\
\bottomrule
\end{tabular}
\end{adjustbox}
\caption{TempTabQA-C Original vs. CounterFactual accuracy and token usage with all baselines.}
\label{tab:appendix-temptabqac-original-counterfactual-all}
\end{table*}

\begin{table*}[!t]
\centering
\small
\setlength{\tabcolsep}{11pt}

\begin{adjustbox}{max width=\linewidth}
\begin{tabular}{l l c c c c}
\toprule
\multirow{2}{*}{Model} & \multirow{2}{*}{Method}
& \multicolumn{2}{c}{Small}
& \multicolumn{2}{c}{Large} \\
\cmidrule(lr){3-4} \cmidrule(lr){5-6}
& & REMS/EMS & TC & REMS/EMS & TC \\
\midrule
GPT-5-nano & Direct Static & 80.10 / 78.71 & 2539 & 60.90 / 58.74 & 4614 \\
 & CLEAR & 56.34 / 55.44 & 4599 & 46.80 / 45.17 & 7423 \\
 & CoT & 66.77 / 65.38 & 2327 & 50.99 / 49.26 & 4722 \\
 & PlanSolve & 68.85 / 67.49 & 2238 & 53.18 / 51.49 & 4661 \\
 & PoT & 6.20 / 6.08 & 3464 & 1.49 / 1.49 & 6674 \\
 & Faithful-CoT & 37.20 / 36.02 & 3311 & 15.43 / 15.06 & 6717 \\
 & SQL Static & 75.87 / 75.79 & 4334 & 66.81 / 66.54 & 4505 \\
\midrule
Llama-70B & Direct Static & 77.56 / 74.39 & 1398 & 55.13 / 50.93 & 2004 \\
 & CLEAR & 68.87 / 67.37 & 2107 & 46.76 / 44.98 & 29164 \\
 & CoT & 83.27 / 81.52 & 1265 & 59.12 / 57.06 & 20654 \\
 & PlanSolve & 81.08 / 79.88 & 11925 & 59.61 / 57.81 & 1964 \\
 & PoT & 6.78 / 6.78 & 1349 & 6.88 / 6.32 & 2399 \\
 & Faithful-CoT & 8.30 / 8.07 & 1445 & 7.43 / 6.88 & 2465 \\
 & SQL Static & 70.39 / 69.79 & 2974 & 65.38 / 64.87 & 2977 \\
\bottomrule
\end{tabular}
\end{adjustbox}
\caption{TempTabQA-C small-table vs. large-table accuracy and token usage with all baselines.}
\label{tab:appendix-temptabqac-small-large-all}
\end{table*}


\subsubsection{Failure Rescue and Qualitative Cases}
\label{sec:qualitative-failure-cases}

The error categories in Figure~\ref{fig:failure_rescue} show which SQL Static
failures are rescued by \textsc{GRIPS}. On TempTabQA-C, the rescued set
includes 24.0\% of SQL Static failures from grounding categories and 14.9\%
from operator categories. On BIRD, the corresponding shares are 11.1\% and
5.2\%. These values describe the audited SQL Static failures rescued by the
complete GRIPS configurations; they do not isolate one Stage~I field.

\begin{figure}[!t]
    \centering
    \includegraphics[width=\linewidth]{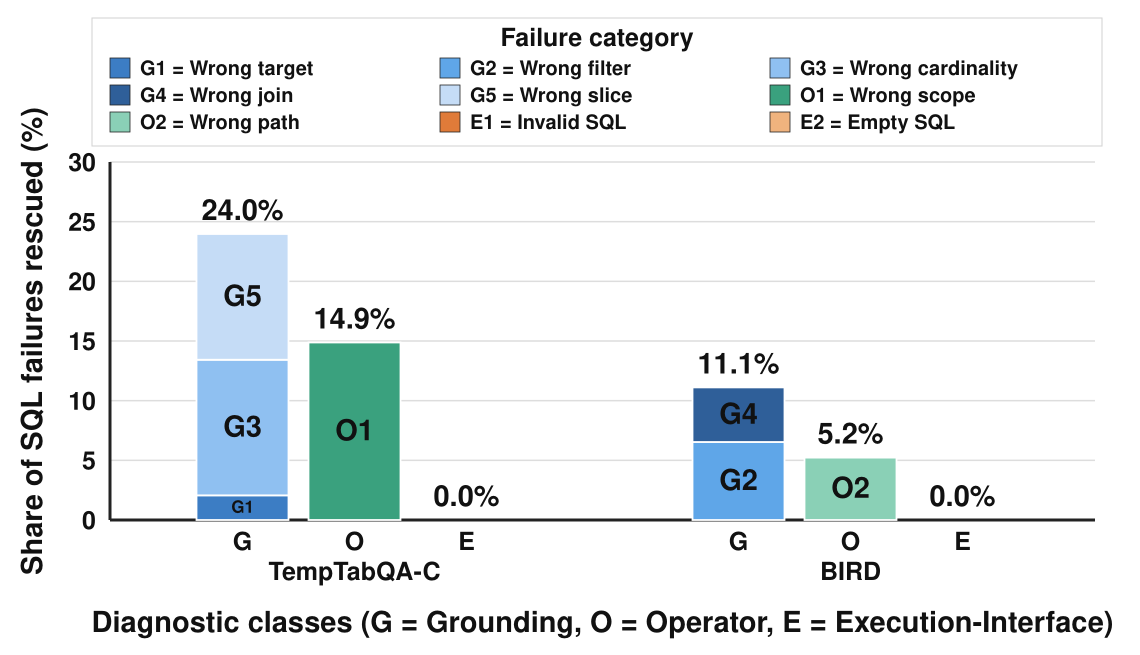}
    \caption{Share of SQL Static failures rescued by \textsc{GRIPS}, by
    diagnostic class and subcategory.}
    \label{fig:app_failure_rescue}
\end{figure}

We complement the aggregate results with deterministically selected cases from
completed full runs. Table~\ref{tab:baseline-failure-examples} uses the
intersection of SQL Static, DIN-SQL, and MAC-SQL failures: the first row is the
smallest question ID uniquely rescued by GRIPS Static, and the second is the
smallest uniquely rescued by GRIPS Adaptive. This shared-target design avoids
comparing unrelated examples while showing that the two GRIPS configurations
can succeed on different cases. Table~\ref{tab:appendix-grips-rescue-examples}
adds TempTabQA-C value and filter errors and BIRD relation and
answer-cardinality errors. Because Adaptive also changes demonstration
selection, these examples do not isolate the effect of Stage~I.

\begin{table*}[!t]
\centering
\small
\setlength{\tabcolsep}{3.4pt}
\begin{adjustbox}{max width=\linewidth}
\begin{tabular}{p{0.25\textwidth}p{0.40\textwidth}p{0.27\textwidth}}
\toprule
\textbf{Question} & \textbf{SQL Static, DIN-SQL, and MAC-SQL failures}
& \textbf{Correct GRIPS outcome} \\
\midrule
\textit{Does Antonio Rebollo have more Silver Medals than Bronze Medals in
Men's teams open?} &
All three baselines attach \textit{Men's teams open} to
\texttt{Format.name} rather than \texttt{Tournament.name}.  SQL Static and
DIN-SQL return the wrong \texttt{Yes}; MAC-SQL returns medal counts instead of
the requested Boolean answer.  All three have EMS/REMS 0/0. &
\textbf{GRIPS Static} filters \texttt{Tournament.name}, compares silver and
bronze counts, and returns \texttt{No}; EMS/REMS are 1/1.  GRIPS Adaptive uses
\texttt{Format.name} and is incorrect. \\
\addlinespace
\textit{Which tournament(s) has Zhang Jike won the most Medals in?} &
SQL Static collapses years by tournament name and returns only
\textit{World Championships}.  DIN-SQL also returns a helper count and keeps
one row; MAC-SQL omits the year and the maximum-count tie filter.  All three
have EMS/REMS 0/0. &
\textbf{GRIPS Adaptive} groups by tournament and year and returns all ten tied
tournament--year events; EMS/REMS are 1/1.  GRIPS Static also collapses the
years and is incorrect. \\
\bottomrule
\end{tabular}
\end{adjustbox}
\caption{Shared-target failures from the completed GPT-5-nano TempTabQA-C
runs.  Among 34 Easy/Original targets on which SQL Static, adapted DIN-SQL, and
adapted MAC-SQL all have EMS $=0$ and at least one GRIPS condition has
EMS $=1$, we select the first Static-only and Adaptive-only success under a
fixed split and source-row order.  The examples are diagnostic rather than
frequency or causal evidence.}
\label{tab:baseline-failure-examples}
\end{table*}

\begin{table*}[!t]
\centering
\small
\setlength{\tabcolsep}{4pt}
\begin{adjustbox}{max width=\linewidth}
\begin{tabular}{p{0.12\linewidth} p{0.24\linewidth}
                p{0.28\linewidth} p{0.28\linewidth}}
\toprule
\textbf{Benchmark} & \textbf{Question}
  & \textbf{GRIPS Static failure}
  & \textbf{Stage-I-augmented Adaptive output} \\
\midrule
TempTabQA-C &
\textit{How many times did Ilya Shymanovich win an
International medal in the European Championships (SC)
before 2019?} &
Splits into \textcolor{BrickRed}{\texttt{t.name =
'European Championships'}} and
\textcolor{BrickRed}{\texttt{f.name = 'SC'}};
adds an incorrect format constraint and returns
\textcolor{BrickRed}{0}. &
Preserves one tournament value
\textcolor{ForestGreen}{\texttt{t.name =
'European Championships (SC)'}};
removes the format filter and recovers
\textcolor{ForestGreen}{count 4}. \\
\addlinespace
TempTabQA-C &
\textit{How many International medals did Jo\~ao Oliva
win in 2019?} &
Invents an \textcolor{BrickRed}{\texttt{International}}
schema filter and applies
\textcolor{BrickRed}{\texttt{t.name LIKE '\%International\%'}}
or location matches; this overconstrains the query and
returns \textcolor{BrickRed}{0}. &
Treats \textit{International} according to the common prompt convention
rather than as a literal schema value; keeps only the recognized
athlete and year constraints and recovers
\textcolor{ForestGreen}{count 1}. \\
\addlinespace
BIRD &
\textit{Name the user that commented
'thank you user93!'} &
Queries \textcolor{BrickRed}{\texttt{UserDisplayName}}
directly from \textcolor{BrickRed}{\texttt{comments}};
the target attribute belongs to the wrong relation, so the
query yields \textcolor{BrickRed}{no answer}. &
Selects the target as a user attribute and uses the join
path \textcolor{ForestGreen}{\texttt{C.UserId = U.Id}};
selects \textcolor{ForestGreen}{\texttt{U.DisplayName}} and
returns \textcolor{ForestGreen}{\textit{Preets}}. \\
\addlinespace
BIRD &
\textit{What is the display name of the user who acquired
the first Autobiographer badge?} &
Matches the earliest badge date correctly, but returns
\textcolor{BrickRed}{all tied users} for that date; this is a
wrong answer-cardinality failure rather than a syntax error. &
Preserves the same join path, but adds an explicit
\textcolor{ForestGreen}{single-answer constraint}
(\textcolor{ForestGreen}{\texttt{ORDER BY}} +
\textcolor{ForestGreen}{\texttt{LIMIT 1}}) and returns
\textcolor{ForestGreen}{\textit{Geoff Dalgas}}. \\
\bottomrule
\end{tabular}
\end{adjustbox}
\caption{Representative paired outputs from real TempTabQA-C and BIRD
instances. \textcolor{BrickRed}{Red} marks the decisive GRIPS Static error;
\textcolor{ForestGreen}{green} marks the corresponding decision in the correct
Stage-I-augmented Adaptive output.  Because Adaptive also changes
demonstrations, these examples illustrate outcome differences rather than
isolated causal effects of the Stage~I handoff.}
\label{tab:appendix-grips-rescue-examples}
\end{table*}

\subsection{Controlled-Handoff Details}
\label{sec:ablation-details}

The controlled comparisons keep the SQL generator, original input,
demonstrations, and generation settings fixed. Only the Stage~I information
changes. Table~\ref{tab:controlled-condition-examples} defines every
condition before the detailed results.

\begin{table*}[!t]
\centering
\footnotesize
\setlength{\tabcolsep}{4pt}
\renewcommand{\arraystretch}{1.18}
\begin{tabularx}{\textwidth}{@{}
>{\raggedright\arraybackslash}p{0.08\textwidth}
>{\raggedright\arraybackslash}p{0.21\textwidth}
>{\raggedright\arraybackslash}X
>{\raggedright\arraybackslash}p{0.27\textwidth}@{}}
\toprule
\textbf{Control} & \textbf{Condition} & \textbf{Definition}
& \textbf{Illustrative example} \\
\midrule
Shared & Without Stage~I information
& The auxiliary Stage~I block is empty. The question, schema/evidence,
demonstrations, and SQL-generation instructions remain unchanged.
& Empty payload: \texttt{\{\}}. \\
Shared & With guided representation
& Supplies recognized question--input matches with their semantic types and
normalized values. No profile summaries are included.
& Column: \texttt{element}\newline Tables: \texttt{bond}, \texttt{atom}. \\
\midrule
Content & With representation + profile
& Adds deterministic counts, structural scope, priority, target determinacy,
and coarse difficulty summaries. This is the deployed GRIPS handoff.
& Representation: \texttt{element}, \texttt{bond}, \texttt{atom}\newline
Implicated tables: 2\newline FK span: 2\newline
Target: clear; difficulty: moderate. \\
Content & With expanded profile
& Adds internal pattern, count/repetition, requirement, and
compressed-signature fields to the deployed handoff.
& Pattern: column over two tables\newline Repeated table mentions: 4\newline
Requirement: 5-tuple\newline Signature: repeated-table doublet. \\
\midrule
Source & With adapted DIN links
& Supplies schema links produced by the DIN-SQL linking stage to the common
Stage~II generator. This is an adapted intermediate output, not native
end-to-end DIN-SQL.
& \texttt{connected.bond\_id = TR004\_8\_9}\newline
\texttt{connected.atom\_id = atom.atom\_id}\newline
Selected column: \texttt{atom.element}. \\
Source & With adapted MAC selection
& Supplies the MAC-SQL selector's table and column decisions to the common
Stage~II generator. The base schema remains available.
& Keep: \texttt{atom}, \texttt{bond}\newline
Select from \texttt{connected}: bond and atom IDs\newline
Drop: \texttt{molecule}. \\
Source & With representation + verified FK
& Supplies the guided representation plus deterministic foreign-key paths
verified against the provided schema. The deployed profile is absent.
& Representation: \texttt{element}, \texttt{bond}, \texttt{atom}\newline
Verified path: \texttt{bond} $\leftarrow$ \texttt{connected}
$\rightarrow$ \texttt{atom}. \\
\bottomrule
\end{tabularx}
\caption{Reader-facing definitions and examples of the controlled Stage~I
information conditions. Examples are condensed from the BIRD question
``What elements are in the TR004\_8\_9 bond atoms?'' The first two conditions
are shared across comparisons. Content conditions cumulatively add
information, whereas source conditions substitute alternative Stage~I
information under the same Stage~II generator.}
\label{tab:controlled-condition-examples}
\end{table*}

\begin{table*}[!t]
\centering
\small
\setlength{\tabcolsep}{3.7pt}

\resizebox{\textwidth}{!}{%
\begin{adjustbox}{max width=\linewidth}
\begin{tabular}{lcccc}
\toprule
\textbf{Stage~II information} & \textbf{TQ--GPT}
& \textbf{TQ--Llama} & \textbf{BIRD--GPT} & \textbf{BIRD--Llama} \\
\midrule
Without Stage~I information & 3798/11.83/86/\$4.09 & 1611/13.02/90/\$0.88 & 3557/13.54/9/\$0.25 & 1170/10.01/24/\$0.04 \\
Guided representation & 3753/11.64/63/\$3.99 & 1636/16.54/69/\$0.89 & 3685/13.06/10/\$0.26 & 1222/12.79/27/\$0.04 \\
Representation + profile & 3999/11.30/106/\$4.21 & 1728/16.20/83/\$0.94 & 3982/12.60/10/\$0.25 & 1579/10.03/29/\$0.05 \\
Expanded profile & 4257/12.31/117/\$4.28 & 1925/17.39/95/\$1.03 & 4402/12.42/11/\$0.27 & 1766/9.90/38/\$0.05 \\
Adapted DIN links & 3874/12.24/76/\$4.10 & 1680/10.91/97/\$0.91 & 3656/12.70/8/\$0.26 & 1226/10.50/24/\$0.04 \\
Adapted MAC selection & 3866/12.16/59/\$4.08 & 1648/10.09/98/\$0.90 & 3849/12.80/7/\$0.27 & 1229/9.20/25/\$0.05 \\
Representation + verified FK & 3802/10.47/76/\$3.95 & 1670/10.34/87/\$0.91 & 3575/12.05/5/\$0.25 & 1246/8.04/26/\$0.04 \\
\bottomrule
\end{tabular}
\end{adjustbox}%
}
\caption{Controlled Stage~II efficiency and reliability. Each cell is mean
total tokens/mean seconds/SQL errors/total USD. Lower is better.}
\label{tab:ablation-efficiency-reliability}
\end{table*}

\begin{table*}[!t]
\centering
\small
\setlength{\tabcolsep}{2.2pt}

\begin{adjustbox}{max width=\linewidth}
\begin{tabular}{llcccc}
\toprule
\textbf{Comparison} & \textbf{Stage~II information}
& \textbf{TempTabQA-C--GPT} & \textbf{TempTabQA-C--Llama}
& \textbf{BIRD--GPT} & \textbf{BIRD--Llama} \\
\midrule
Shared & Without Stage~I information
& 71.67 & 75.06 & 39.09 & 36.59 \\
Shared & With guided representation
& 74.94 & 77.10 & 38.77 & \textbf{39.22} \\
\midrule
Content & With representation + profile
& \textbf{75.52} & \textbf{77.86} & \textbf{40.34} & 38.41 \\
Content & With expanded profile
& 74.08 & 76.57 & 37.13 & 35.14 \\
\midrule
Source & With adapted DIN links
& 73.01 & 75.22 & 38.04 & 37.32 \\
Source & With adapted MAC selection
& 73.12 & 75.71 & 37.22 & 37.32 \\
Source & With representation + verified FK
& 75.19 & 76.73 & 37.58 & 35.87 \\
\bottomrule
\end{tabular}
\end{adjustbox}
\caption{Parity-controlled Stage~II handoff comparison
(accuracy \%; higher is better). Bold indicates the highest observed accuracy
within each benchmark--model block.}
\label{tab:full-controlled-ablation}
\label{tab:controlled-grounding}
\end{table*}



For arm $a$, executed answer $\hat y_i^{(a)}$, gold answer $y_i$, and the
benchmark canonicalizer $M$, we compute
\begin{equation}
\mathrm{EMS}(a)=\frac{100}{N}\sum_i
\mathbb{1}\!\left[M(\hat y_i^{(a)},y_i)=1\right].
\end{equation}
REMS gives a multi-item answer the fraction of gold items recovered; exact
canonical scalar matches receive one and other scalar answers receive zero.
TempTabQA-C Strict EMS removes the scalar-containment allowance and
requires canonical scalar equality or exact set equality.  Empty outputs and
non-executable SQL remain in the denominator and score zero.  For paired EMS
contrasts we use 10,000 deterministic target-level bootstrap resamples and an
exact two-sided McNemar test.  The $p$-values below are unadjusted because the
transitions were pre-specified component contrasts rather than a family used
for model selection.

Table~\ref{tab:ablation-efficiency-reliability} reports only the controlled
Stage~II model call.  Tokens equal provider-reported input plus output tokens;
latency is elapsed model-call time.  Costs use provider-reported values or the
recorded price snapshot.  The table excludes deterministic preprocessing,
retrieved-artifact construction, database execution, and evaluation.  The
adapted DIN/MAC conditions additionally exclude the upstream cost of producing
their precomputed Stage~I artifacts.

\section{AI Disclosure}
\label{sec:ai-disclosure}

AI-based assistants were used for limited writing, coding, and
figure-preparation support during the development of this project. All
experiments, annotations, technical claims, and final manuscript decisions
were reviewed and validated by the authors.


\label{sec:appendix}

\end{document}